\documentclass[journal]{IEEEtran}
\usepackage{times}
\usepackage{epsfig}
\usepackage{graphicx}
\usepackage{amsmath}
\usepackage{amssymb}
\usepackage{makecell}
\usepackage{caption}
\begin{document}
%
\title{Monocular Depth Estimation with Directional Consistency by Deep Networks}
%
%
%

\author{\IEEEauthorblockN{Fabian Truetsch, Alfred Sch\"ottl\\}
\IEEEauthorblockA{University of Applied Sciences Munich, Department of Electrical Engineering and Information Technology}}
%
%

\markboth{}%
{}
%



\maketitle

\begin{abstract}
As processing power has become more available, more human-like artificial intelligences are created to solve image processing tasks that we are inherently good at. As such we propose a model that estimates depth from a monocular image.
Our approach utilizes a combination of structure from motion and stereo disparity. We estimate a pose between the source image and a different viewpoint and a dense depth map and use a simple transformation to reconstruct the image seen from said viewpoint. We can then use the real image at that viewpoint to act as supervision to train out model. The metric chosen for image comparison employs standard L1 and structural similarity and a consistency constraint between depth maps as well as smoothness constraint.
We show that similar to human perception utilizing the correlation within the provided data by two different approaches increases the accuracy and outperforms the individual components.\end{abstract}


%
\IEEEpeerreviewmaketitle

\section{Introduction}
\begin{figure}
\includegraphics[width=\linewidth]{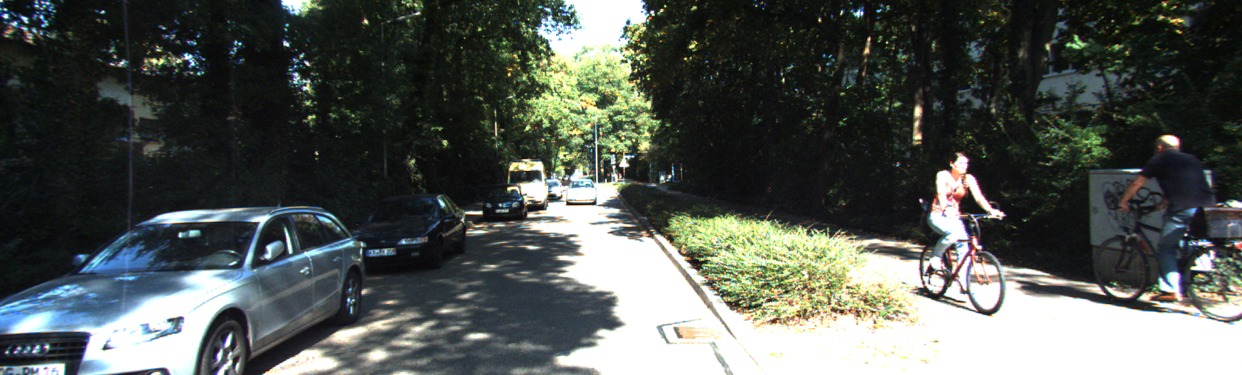} \\
\includegraphics[width=\linewidth]{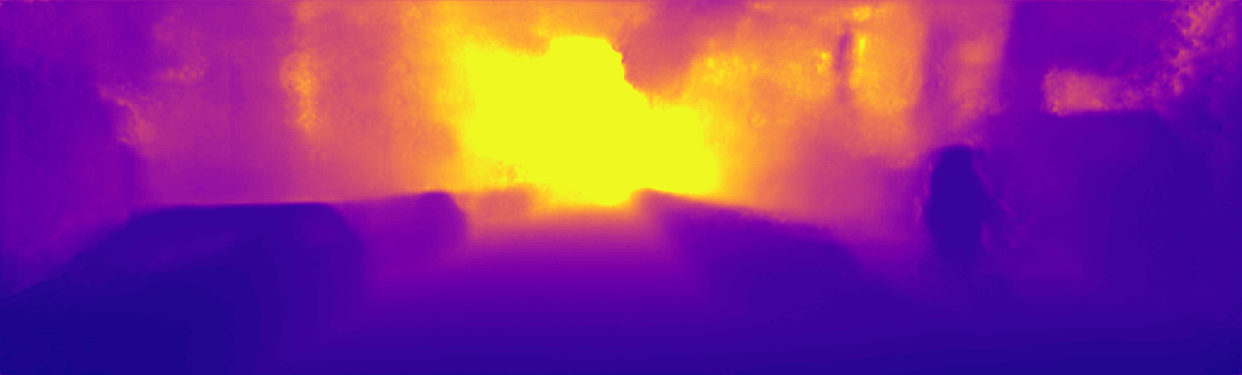}
\caption{Shown is a depth map. The pixel value represents the distance from the camera.}
\label{fig:depth}
\end{figure}

It has been shown that image processing utilizing neural networks has surpassed classical approaches on general tasks like classification  \cite{krizhevsky_imagenet_2017}, detection of objects  \cite{zeng_crafting_2016} and generation of images  \cite{gong_motion_2016}. This project focuses on image generation, more precisely generating a depth map (as shown in \ref{fig:depth}) from a monocular image. \\
Generating depth maps is a discipline with a long history of different approaches using methods from two groups. Active devices include sonars, emitting auditive signals and listening to its reflection  \cite{glisson_sonar_1970}, structured light, projecting a light pattern onto the scene and extracting depth from change of its shape \cite{boyer_color-encoded_1987}, and LIDAR, finding the depth by measuring the time between an emitted impulse and its reflection \cite{goyer_laser_1963}. Passive devices include stereo cameras, inferring depth by finding the disparities between two images \cite{induchoodan_depth_2014}, or modifying the focal length of a single camera to find the optimal sharpness for each pixel \cite{xiong_depth_1993}. While they have competitive accuracy, each of these approaches have disadvantages like being expensive, having problems with ambient light \cite{gupta_structured_2013} or being inaccurate in areas of low texture \cite{sach_robust_2009}. 
To combine high accuracy, low cost and robustness to environmental factors, researchers have turned to machine learning algorithms utilizing supervised learning on big amounts of ground truth data to infer depth onto an image taken by a single camera \cite{eigen_depth_2014}. \\
Historically, acquisition of ground truth depth data has proven to be quite a challenge, even more so than on other machine learning tasks \cite{haltakov_framework_2013}. Consequently current approaches circumvent this issue by posing the problem as an image reconstruction task. In this work the approach of reconstructing a stereo image \cite{godard_unsupervised_2016} and estimating a different viewpoint similar to a structure by motion approach \cite{zhou_unsupervised_2017} is combined.

\section{Related work}

The proposed model consists of two major components. The pose estimator yields an approximate relative camera pose between two training images. The depth estimator gives a dense depth map of an image. By combining these two parts the need for supervised input data can be eliminated. The depth estimator is mainly inspired by the stereo disparity approach of Godard et al. \cite{godard_unsupervised_2016} while the pose estimator is loosely related to the structure from motion approach of Zhou et al. \cite{zhou_unsupervised_2017}.

\subsection{Structure from motion}
Estimating structure from motion (SfM) has a range of established pipelines using sequential data of close spatial proximity \cite{mur-artal_orb-slam:_2015} or global pipelines, estimating the pose of all cameras in a first step and calculating the structure afterwards \cite{moulon_global_2013}.

\subsection{Warping based view synthesis}
Synthesizing a scene from a novel camera viewpoint is the counterpart to SfM. First the underlying structure has to be understood, e.g. by pixel correspondences from different input views, then a new viewpoint can be synthesized by composing image patches from the input \cite{chen_view_1993}, \cite{fitzgibbon_image-based_2003}. To be effectively used in neural networks, the transformation has to be differentiable, giving rise to spatial transformer networks \cite{jaderberg_spatial_2015}.

\subsection{Supervised depth estimation}
With the rise of machine learning different approaches to infer depth into a single 2D image were developed. Saxena et al. proposed Make3D, segmenting images into planes and predicting their orientation and location within the surrounding patches. The group uses a Markov random field and train on a set of images with depth data from laser scans, achieving a good general estimation but having problems with small structures and having no global context to refer upon \cite{saxena_make3d:_2009}. Using neither segmentation nor handcrafted features Eigen et al. created a convolutional neural network (CNN) that estimates a coarse depth map of the scene and then adds detail it in a second set on convolutions \cite{eigen_depth_2014}. Later they improved upon their approach by adding a third scale of convolutions and refining their network architecture, showing that by applying small changes in the output and loss function the model can be used, not just for depth estimation, but also to predict surface normals and semantic labels \cite{eigen_predicting_2014}.

\subsection{Unsupervised depth estimation}
While the supervised models had great success, it is impractical to acquire  large amounts on depth annotated training data. Thus researchers turned to unsupervised learning aiming for a metric, that allows them to accurately describe the quality of an estimated depth map. \\
The choice fell on a metric that has been in use to quantify the error of classic stereo algorithms. Proposed by Szeliski the disparity is evaluated by synthesizing a novel viewpoint from the input images and the disparities and comparing it to a real picture \cite{szeliski_prediction_1999}. The error metric was not the focus of his work, hence an improvement can be  achieved by choosing an appropriate method to compare images. Recent work shows that a combination of a linear error and structural similarity (SSIM \cite{wang_image_2004}) is suited very well for training in neural networks, resulting in a reduced error and objectively higher quality images \cite{zhao_loss_2015}. \\
Deep3D by Xie et al.  use stereo images for training and creates a depth map by producing a probability distribution over all possible disparities for each pixel, synthesizing the right view from the most probable pixels of the left image \cite{xie_deep3d:_2016}. This approach is limited by exponentially increasing memory consumption of the probability distributions, making it unfeasible for larger images.\\
Godard's approach \cite{godard_unsupervised_2016} infers depth by generating a stereo twin of the input image. They use an autoencoder styled network based on DispNet \cite{mayer_large_2016} for estimating the disparity map. The left camera's image is warped with the disparity map and the distance between the cameras given by the dataset. The resulting image is compared to the right camera's image with SSIM and a linear difference. Additionally they employ a left-right consistency check and gradient smoothness for the disparity map.\\
Zhou et al. \cite{zhou_unsupervised_2017} base their approach on SfM. Using the same autencoder style, their model trains on a series of images by predicting a depth map and in a separate smaller model a 6 DoF camera pose and an explainability mask\footnote{The pixel value in the explainability mask represents the likelihood that the respective depth pixel in the image is estimated correctly. } between pictures in a series. The images get projected onto a target image and are compared with a pixel by pixel linear error scaled with the explainability mask and a gradient smoothness loss for the depth map is applied.

\section{Method}
For training data we rely on the popular KITTI dataset \cite{geiger_are_2012}. A training sample consists of two subsequent frames taken by a stereo camera mounted on a moving car which are denoted by $I_l$ and $I_r$ for the first frame and $ I_{l+1}$ and $I_{r+1}$ for the second. Furthermore the camera calibration matrix $C$  is known. Note that only $I_l$ is given as an input to the depth module while the other three images act as supervision.
\subsection{View synthesis}
\label{sec:warp}
\begin{figure} 
\includegraphics[width=0.9\linewidth]{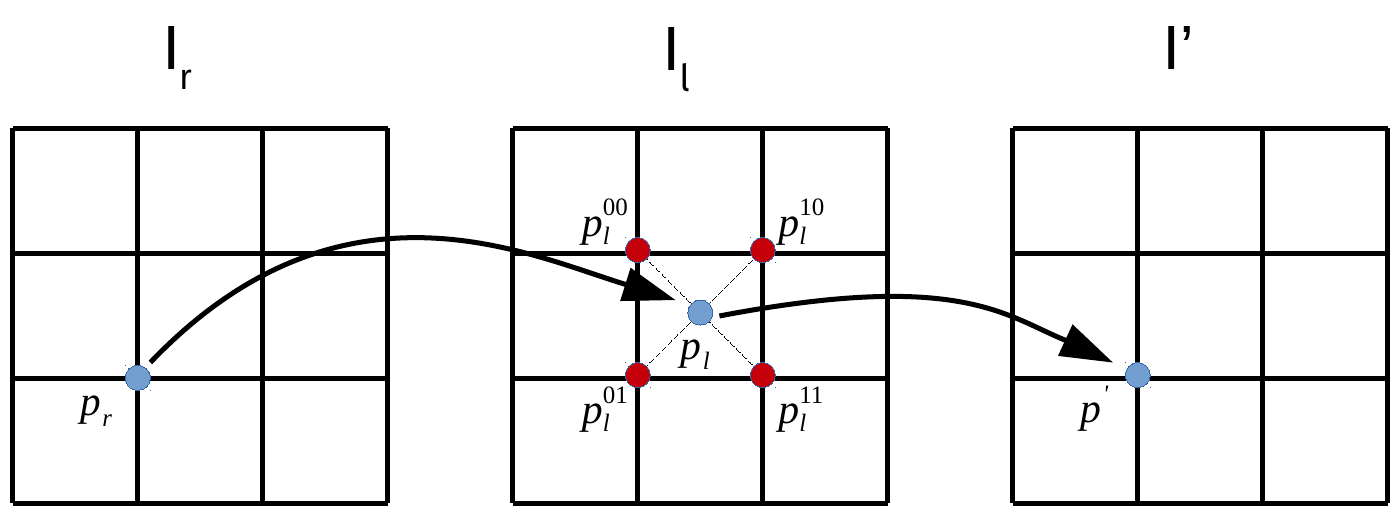} 
\caption{For each point in the target image $I_r$ the corresponding point in the source image is calculated based on the depth and pose. As the projection is continuous the value for $p^{'}$ is interpolated from the surrounding pixels in the source image, $p_l^{xy}$}
\label{fig:sampling}
\end{figure}

Given a depth map $D$ and a pose encoded in a transformation matrix $T_{l}^{r}$ we transform the source image $I_l$ to a target image denoted by $I_r$ to acquire a transformed image $I_l^{'}$.\\
We first transform the homogeneous points from the source sensor $p^{l}$ to points in the world system $p^{w}$ by adding in the estimated depth $d$ from the depth map $D$.
\begin{align}
p^w = 
\begin{pmatrix}
C_l^{-1} | 0
\end{pmatrix}
p^l d 
\end{align}
Afterwards we transform the pixels to the target camera plane utilizing the estimated 6 DoF camera movement $T_{l}^{r}$ and back to the respective sensor points $p^{l'}$ with the intrinsic matrix.
\begin{align}
p^{l'} = 
\begin{pmatrix}
C_{l}|0
\end{pmatrix} 
T_{l}^{r} p^{w} d^{-1}
\end{align}
We shorten the transformation for all points of an image $I$ for further use to
\begin{align}
I' = T(I,D,T_{I}^{r})
\label{eq:transform}
\end{align}
The last step is shown in \ref{fig:sampling}. The pixel values in the transformed image represent the coordinates in the source image. Note that these coordinates are continuous so the RGB value has to be interpolated from the surrounding pixels in the source image.
%

\subsection{Network architecture}
\begin{figure} 
\includegraphics[width=0.9\linewidth]{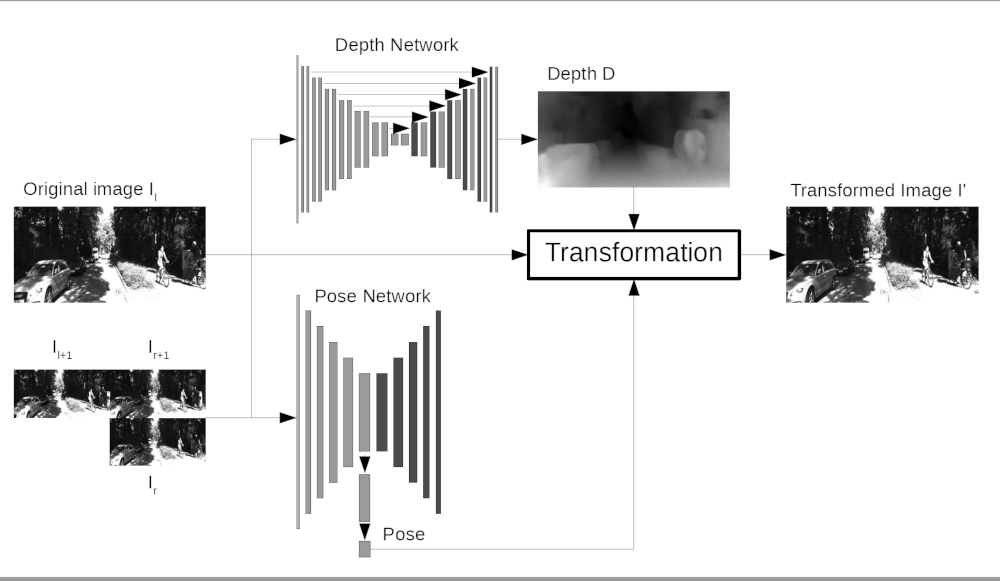} 
\caption{The depth network receives only $I_l$ as an input, while the pose network gets all 4 available images. $I_l$ is then warped as described in \ref{sec:warp} and sampled to receive the transformed image $I_{'}$. }
\label{fig:signal_flow}
\end{figure}

\begin{figure} [!ht]
\includegraphics[width=0.9\linewidth]{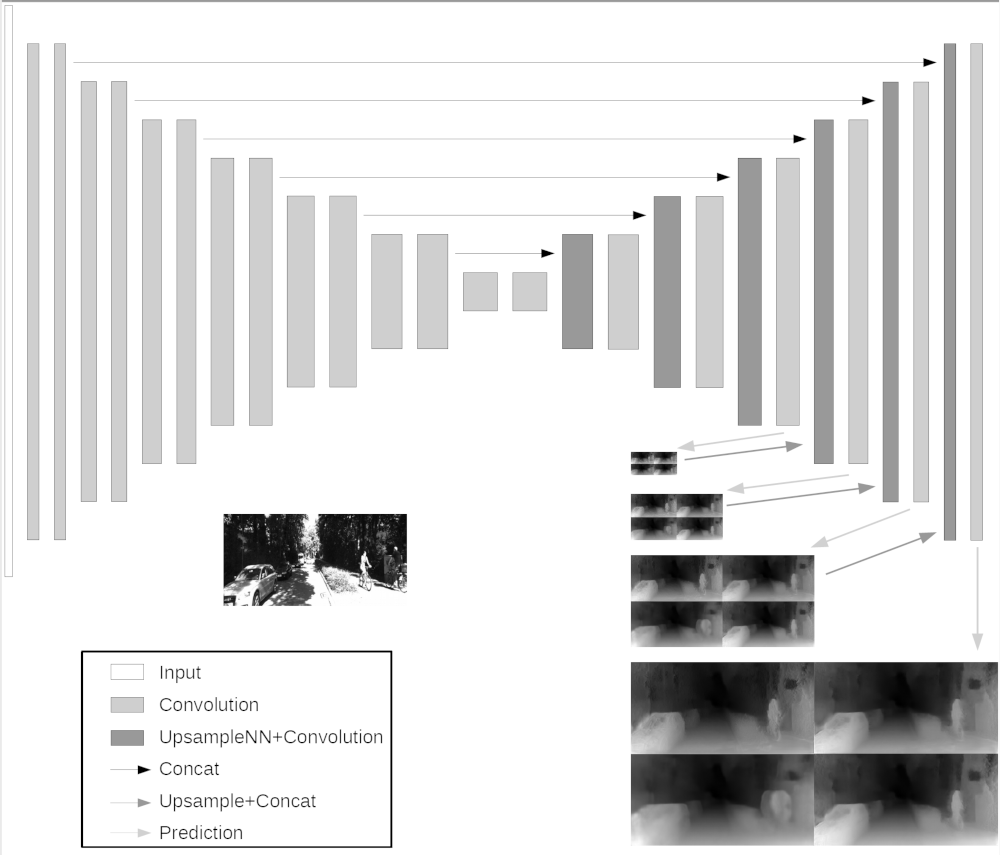}
\caption{The width and height of the blocks indicate the size of the image at that scale and the amount of channels respectively. Each increase or decrease represent a factor of 2. For the input image 4 depth maps are estimated at each of the output scales. The upsampling is done using nearest neighbors.}
\label{fig:vgg}
\end{figure}

The input of the depth module is the first image from the left camera, $I_l$. $I_r$, as well as $I_{l+1}$ and $I_{r+1}$ are used for supervision. The pose module receives all four images as an input.\\
A general overview of the process can be found in Figure \ref{fig:signal_flow}. Four depth maps and two poses, from $I_l$ to $I_r$ and from $I_l$ to $I_{l+1}$ represented by the transformations $T_l^r$ and $T_l^{l+1}$ are estimated in their respective modules and used to transform the input image. The loss to train the network is composed of metrics between transformed images to real ones. Note that the pose network is only used for the training and the depth inference works on its own once the network is fully trained.\\
The depth network's architecture (refer to figure \ref{fig:vgg}) is a U-shaped network (see also \cite{godard_unsupervised_2016}) utilizing an autoencoder design with skip connections and a multi-scale output to estimate the depth map.
The image $I_l$ serves as the only input but returns a set of 4 depth maps, one for each image used during training. As an activation function exponential linear units (Elu) \cite{arn_elu_2015} are used as they increase the speed of convergence and the accuracy. Upsampling is done by a classic nearest neighbour approach coupled with a convolution.

\begin{figure} [!ht]
\includegraphics[width=0.9\linewidth]{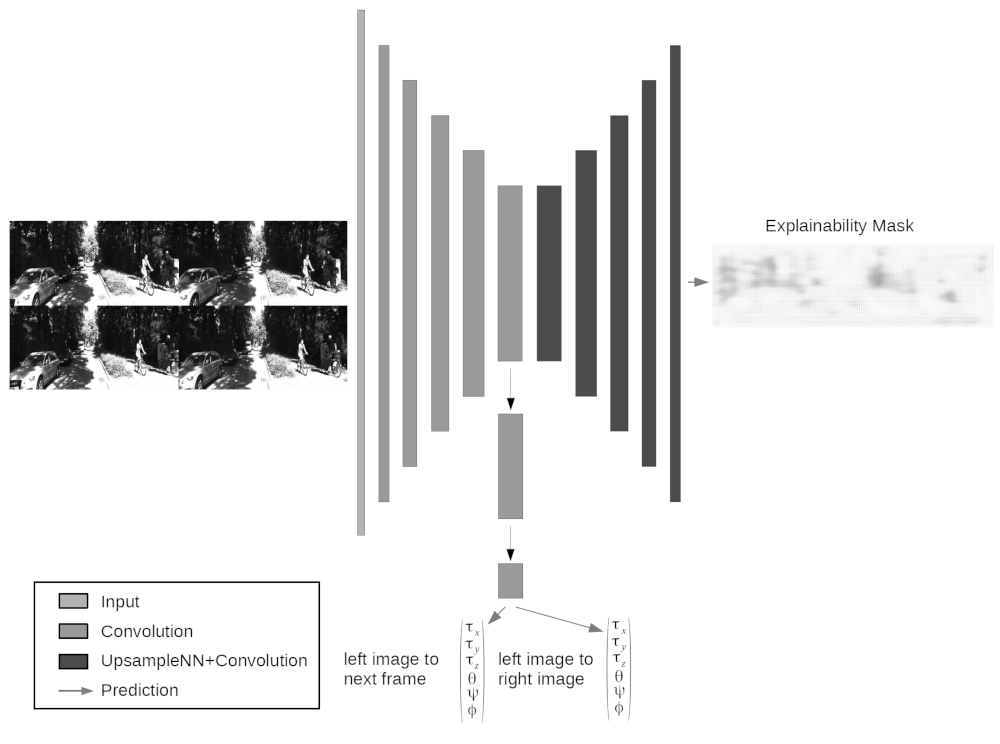}
\caption{Shown is the network architecture for the pose- and explainability mask estimation module. The width and height of the blocks indicate the size of the image at that scale and the amount of channels respectively. Each increase or decrease represent a factor of 2. For each set of 4 input images 1 explainabilty mask and 2 sets of 6 DoF pose estimations are returned. White patches in the explainability mask represent areas where the network is confident about the estimated depth. The 2 sets of poses represent the camera movement between frames and the difference between the left and right camera.}
\label{fig:pose}
\end{figure}

The pose estimation, shown in \ref{fig:pose}, is trained in a separate pose network consisting of convolutions with a stride of 2 to reduce the size and a final 1x1 convolution. It outputs two sets of 6 DoF coded pose estimations, $T_{l}^{l+1}$ and $T_l^r$. They represent the camera pose of the next frame and the right camera respectively relative to $I_l$. As the camera is mounted on a car rotations during a short timeframe are small, therfore the pose of the right camera in the next frame is approximated by the sum of both. 

\subsection{Training loss}
The model's training loss consists of several parts.

\paragraph{Image Loss}
The generated depth map should allow the transformation to calculate the image from a novel camera pose perfectly in all regions that can be observed. The generated image can be compared with the actual image with a combination of a single scale SSIM term and a simple L1 loss \cite{zhao_loss_2015}. The image loss is given as

\begin{align}
l_{image} =& \frac{1}{N} \sum_{i,j} E_{ij}^m \left( \alpha \, l_{SSIM} + (1-\alpha)\, l_{L1} \right)\label{eq:l1}
\end{align}
with N being the number of pixels and $E_{ij}^m$ being the values of the explainability mask discussed in paragraph \ref{par:exp}. $\alpha = 0.85$ is a scaling factor, weighting the loss towards the SSIM metric. $l_{L1}$ is given by
\begin{align}
l_{L1} = \lVert I_{ij}^r - I_{ij}^{'} \rVert_1
\end{align}
while $l_{SSIM}$ is defined as
\begin{align}
l_{SSIM} = \frac{1- SSIM(I_{ij}^ r, I_{ij}^{'})}{2}
\end{align}
SSIM itself is given by
\begin{align}
&SSIM(x,y) = \frac{(2\mu_x\mu_y + c_1)(2\sigma_{xy} + c_2)}{(\mu_x^2 + \mu_y^2 + c_1)(\sigma_x^2 + \sigma_y^2 + c_2)} \label{eq:ssim}\\
\end{align}
It iterates over blocks of size 3x3 with $c_1= 0.01$, $c_2=0.03$.\\ 
This function is applied to all scales and on all generated images. 

\paragraph{Depth Smoothness}
Natural scenes are usually cluttered with objects. A full segmentation of these objects is a very challenging task, but there is a lot to learn even from a partial segmentation into small patches. While depth can or can not change at edges in the image,it is much more likely that there is no depth discontinuity within a small patch than at the edges. 
Therefore this term penalizes non-smooth reconstructions by
\begin{align}
l_{ds} = \frac{1}{N} \sum_{i,j} \lVert \partial_x d_{ij}\rVert_1 e^{-\left|\lVert \partial_x I_{ij} \rVert_1 \right|} + \lVert \partial_y d_{ij}\rVert_1 e^{-\left |\lVert \partial_y I_{ij} \rVert_1 \right|}
\end{align}
with the gradients
\begin{align}
\partial_x d_{ij} = d_{i,j}-d_{i+1,j} \quad \text{and} \quad \partial_y d_{ij} = d_{i,j}-d_{i,j+1}
\end{align}
Note that these gradients are weighted by the negative exponential of the similarly calculated gradients in the real image. 

\paragraph{Directional Consistency}
Stereo methods usually include a left-right consistency check as a post processing step (i.e. in \cite{zbontar_stereo_2015}, \cite{jie_left-right_2018}). Matching pixels in the left and right image of a stereo camera have the same depth. Consequently projecting the depth map of the left image similar to how the image itself is projected, should result in the depth map of the right image and the other way around. Areas with differences in depth can often be found in regions that are occluded in the target image, so additional information about occlusions is gained. Given that the network predicts four depth maps, the consistency check can be extended to the forward motion of the car. The depth map itself is transformed and compared to the other estimated depth maps.
\begin{align}
l_{lr} = \frac{1}{N}  \lVert T(D_l,D_l,T_l^r) - D_r \rVert_1
\end{align}

\paragraph{Explainability Loss}
\label{par:exp}
The input consists of two frames taken at different points in time. This implies that moving objects, like other cars, could have changed their position between those frames. Therefore an explainability mask $E_m$ similar to  \cite{zhou_unsupervised_2017} is implemented. The pixels of the explainability mask represent the probability that the corresponding pixel in the target image can be correctly represented by warping the source image. As the mask is multiplied pixelwise a trivial solution for a global minimum is a mask with only zeros. To counteract the trivial solution a cross-entropy loss with a target of 1 is applied to the explainability mask.
\begin{align}
l_{exp} = \frac{1}{N}\sum_{ij} log(E^m_{ij})
\end{align}

\paragraph{Total Loss}
The total loss function is obtained by bringing the different parts together:
\begin{align}
l_{tot} = l_{image} + l_{ds} + l_{lr} + l_{exp}
\end{align}

\section{Results}
\begin{figure}
Input \hspace{0.14\linewidth}  Zhou et al.\cite{zhou_unsupervised_2017}  Godard et al. \cite{godard_unsupervised_2016} \hspace{0.02\linewidth} Ours \\
\includegraphics[width=0.24\linewidth]{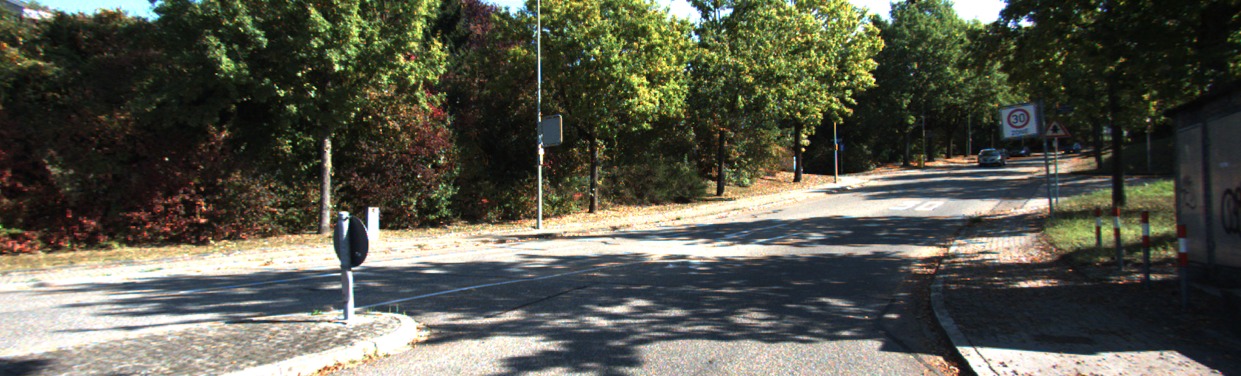} \includegraphics[width=0.24\linewidth]{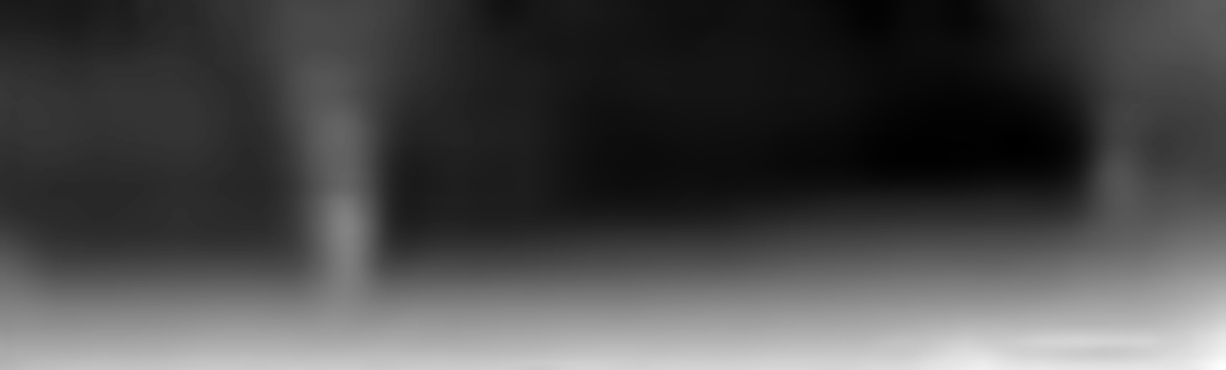} \includegraphics[width=0.24\linewidth]{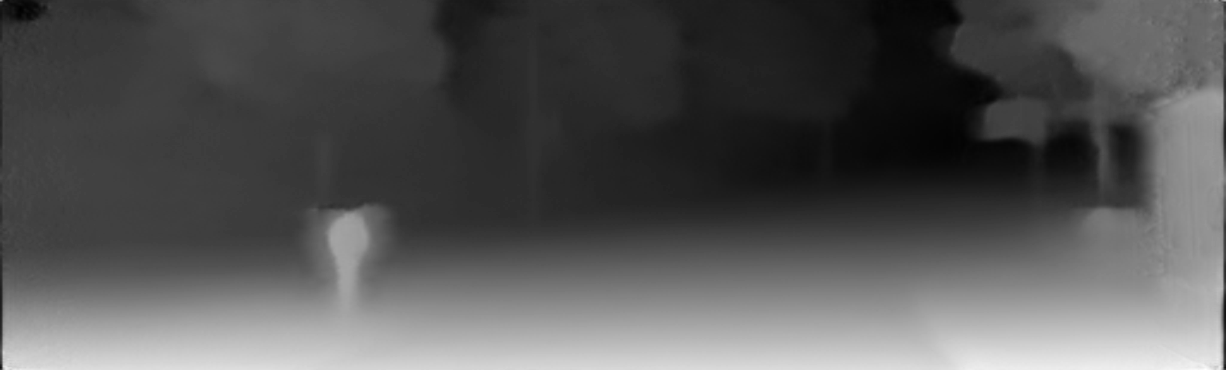} \includegraphics[width=0.24\linewidth]{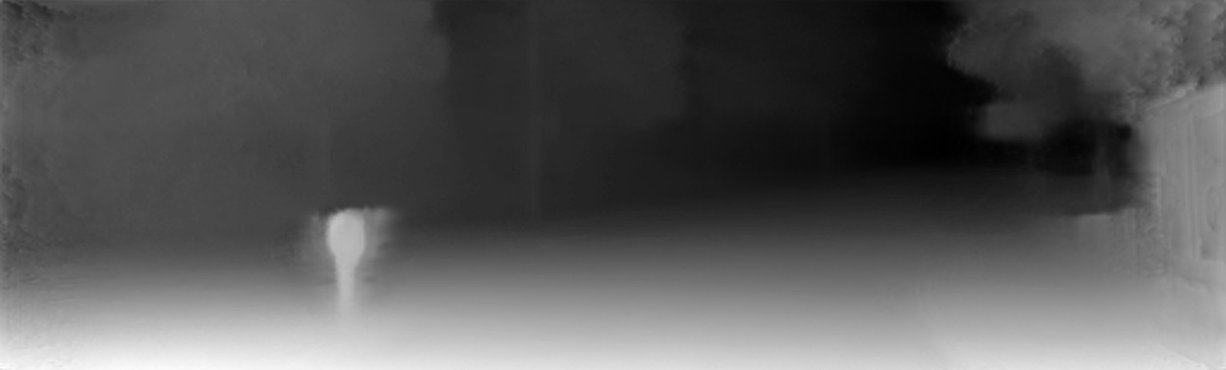} \\
\includegraphics[width=0.24\linewidth]{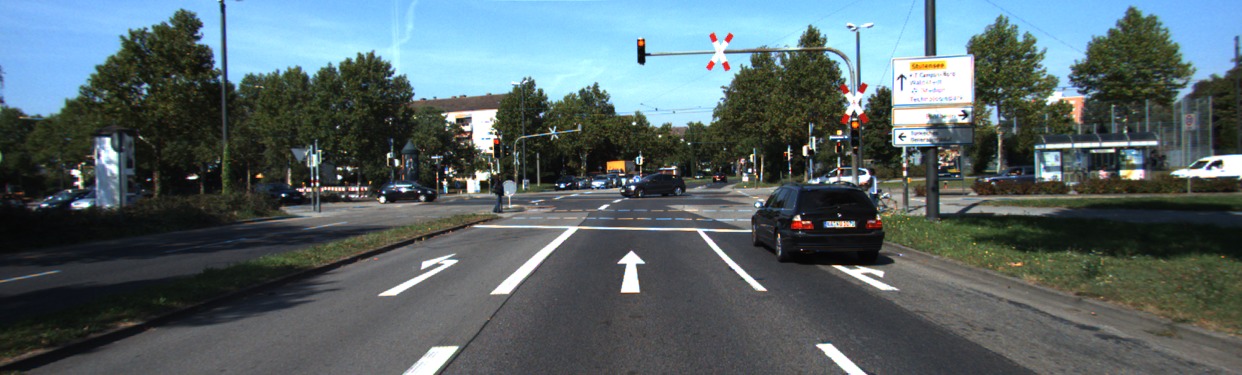} \includegraphics[width=0.24\linewidth]{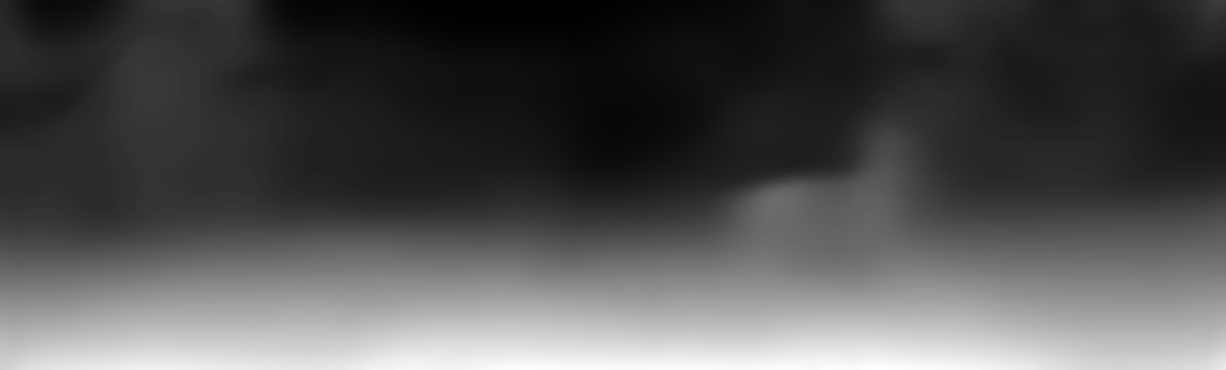} \includegraphics[width=0.24\linewidth]{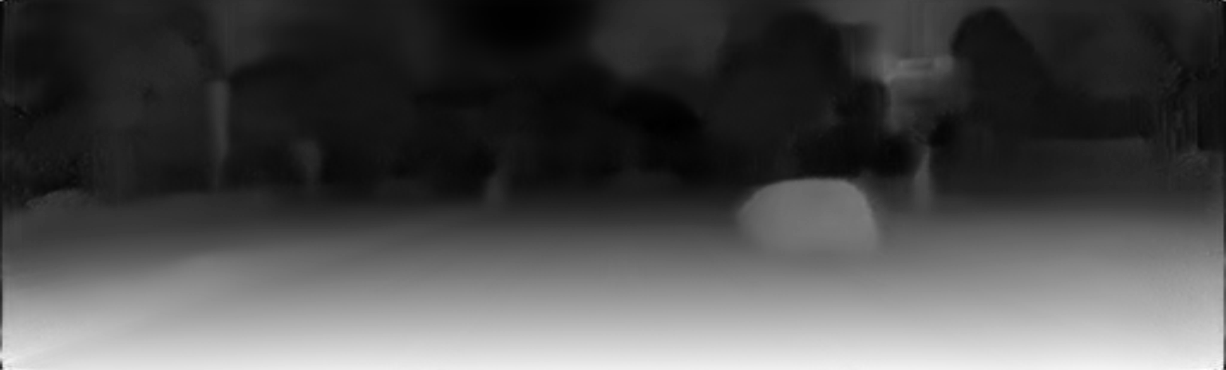} \includegraphics[width=0.24\linewidth]{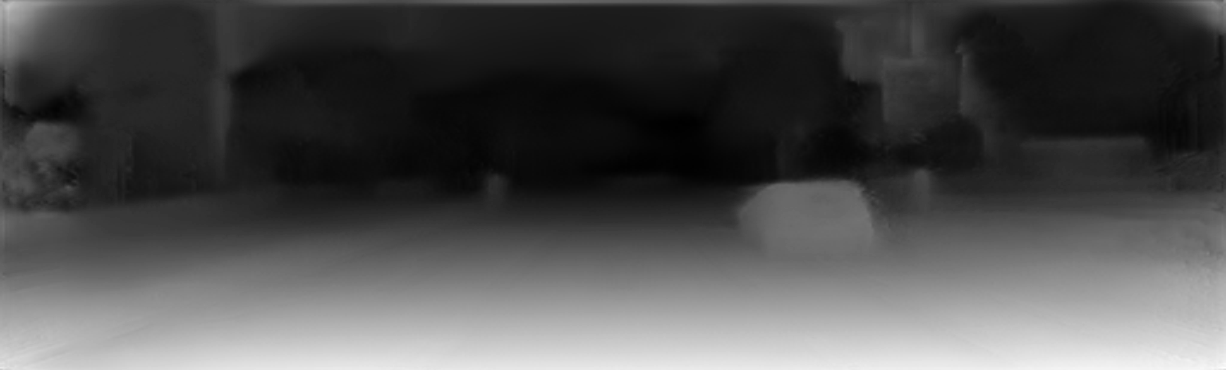} \\
\includegraphics[width=0.24\linewidth]{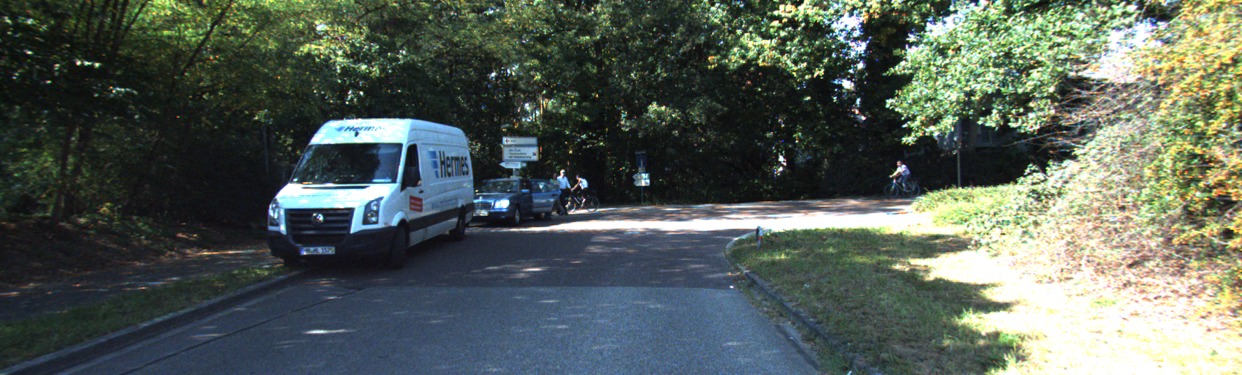} \includegraphics[width=0.24\linewidth]{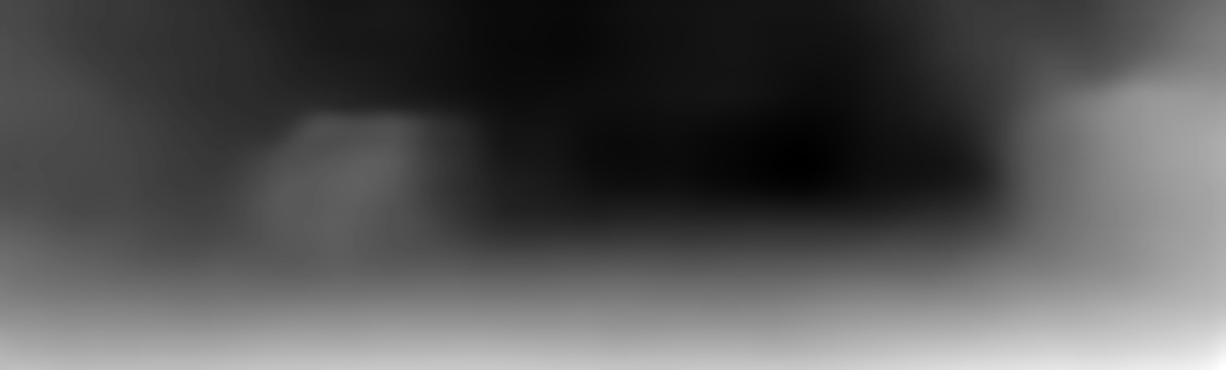} \includegraphics[width=0.24\linewidth]{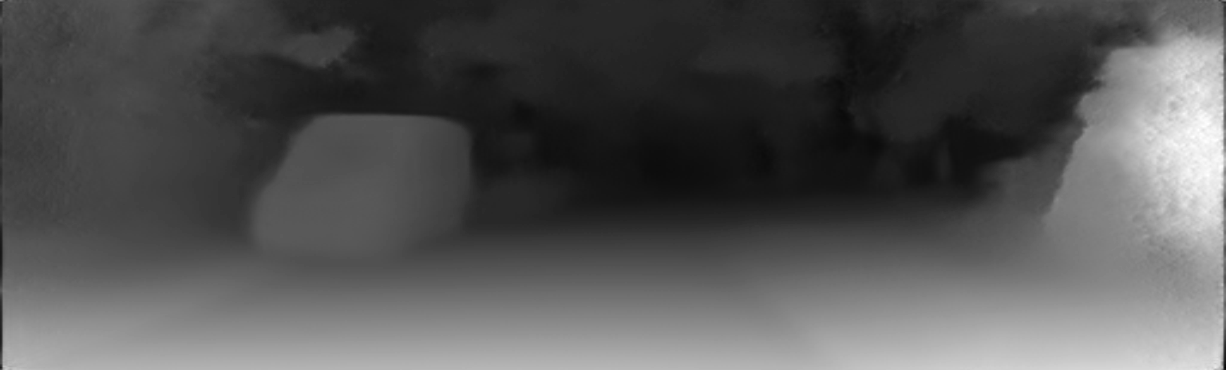} \includegraphics[width=0.24\linewidth]{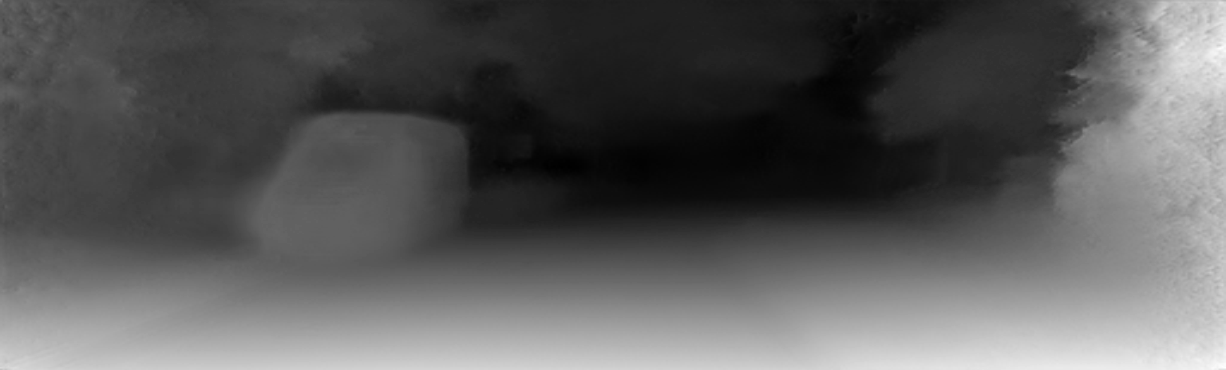}\\
\includegraphics[width=0.24\linewidth]{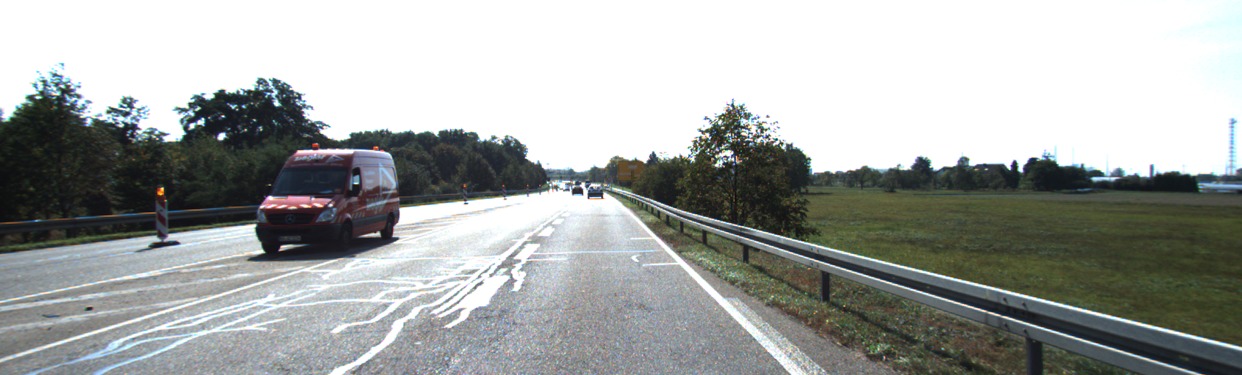} \includegraphics[width=0.24\linewidth]{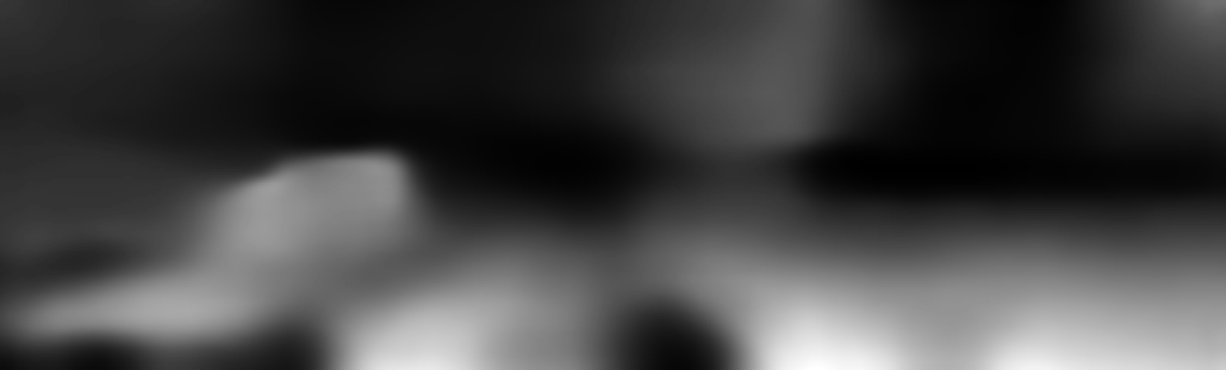} \includegraphics[width=0.24\linewidth]{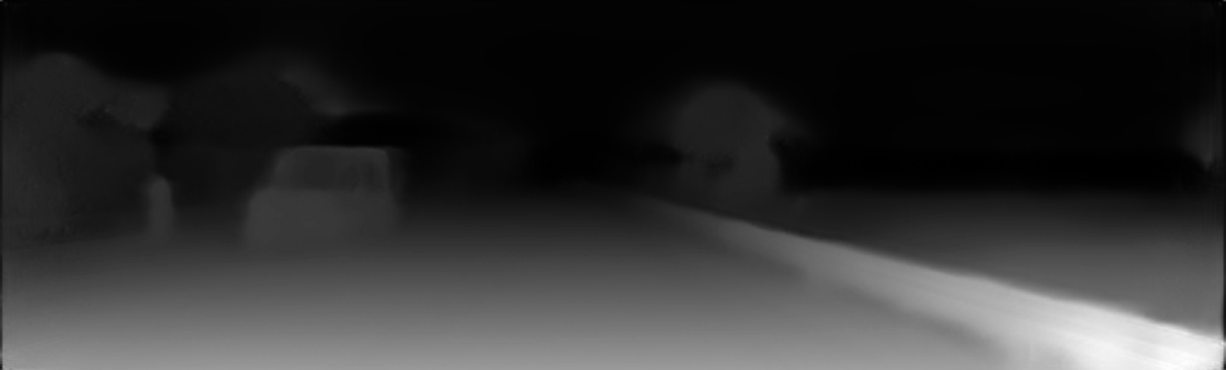} \includegraphics[width=0.24\linewidth]{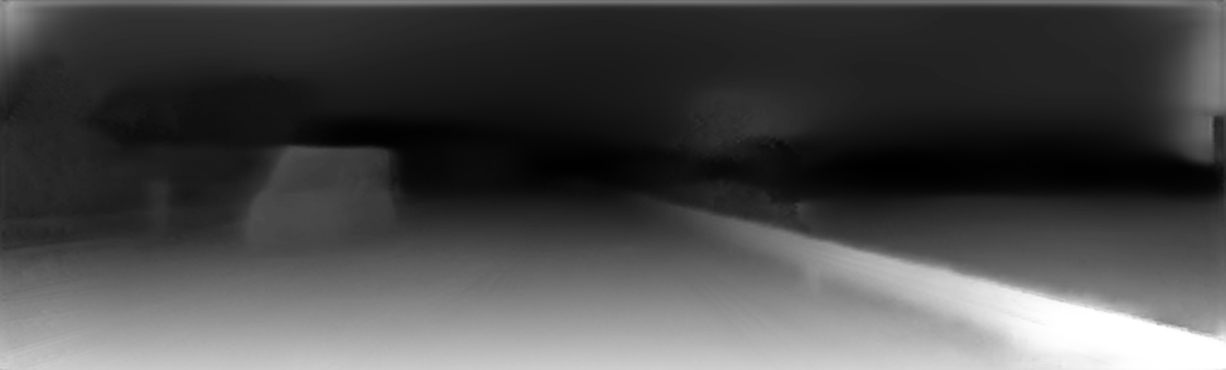}\\
\includegraphics[width=0.24\linewidth]{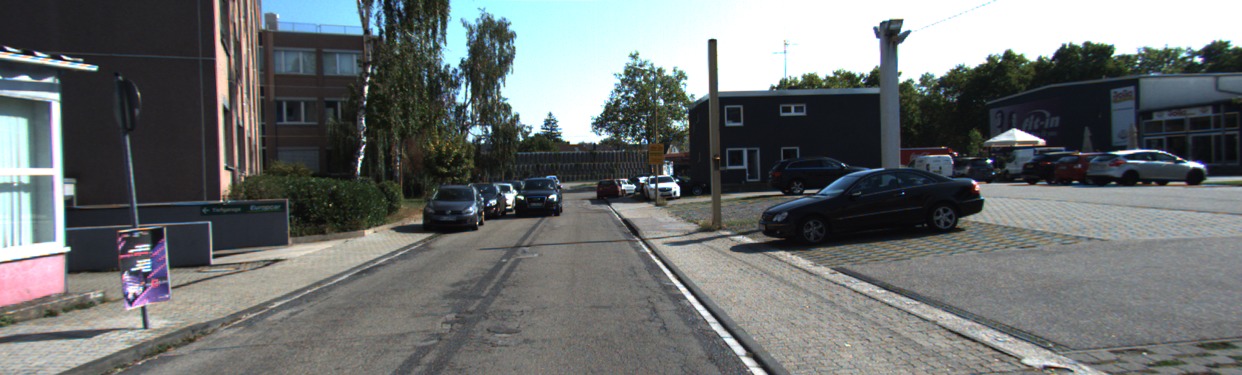} \includegraphics[width=0.24\linewidth]{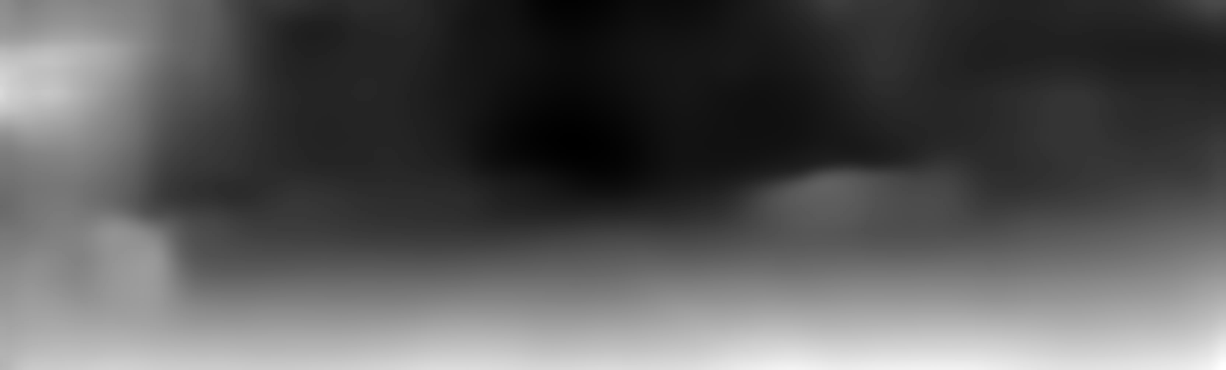} \includegraphics[width=0.24\linewidth]{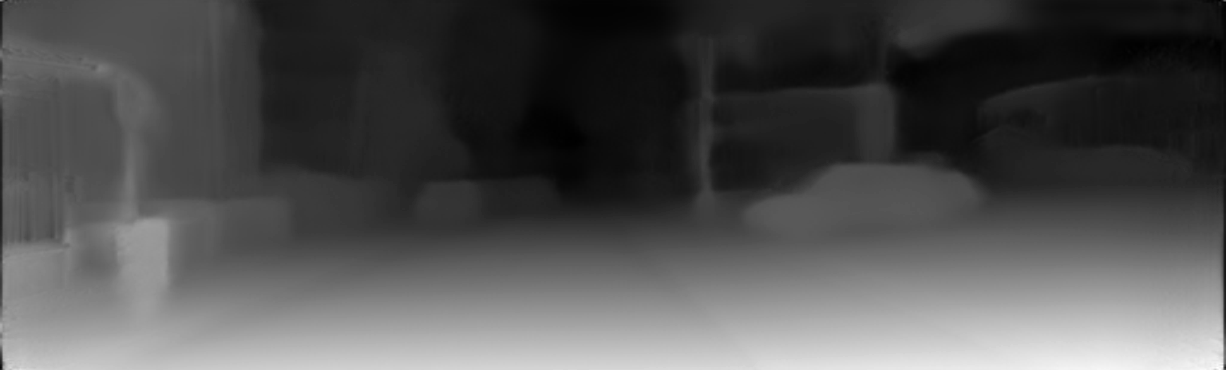} \includegraphics[width=0.24\linewidth]{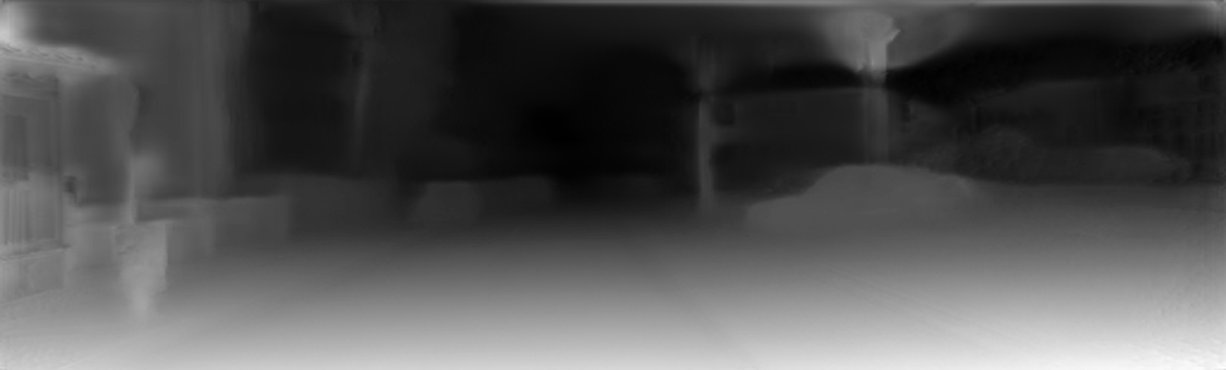}\\
\includegraphics[width=0.24\linewidth]{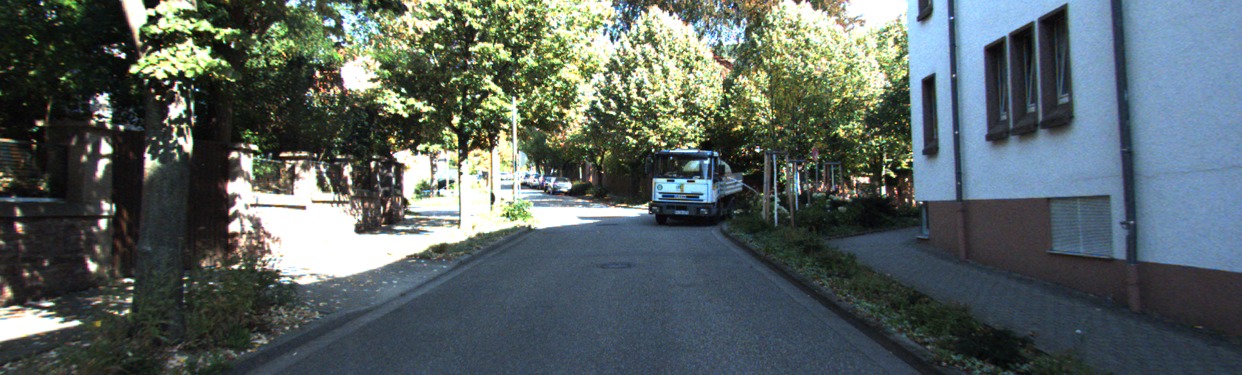} \includegraphics[width=0.24\linewidth]{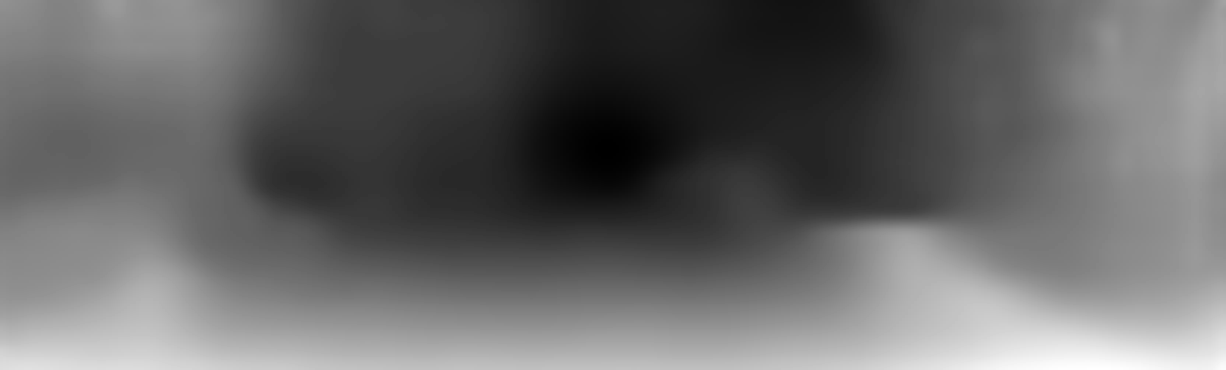} \includegraphics[width=0.24\linewidth]{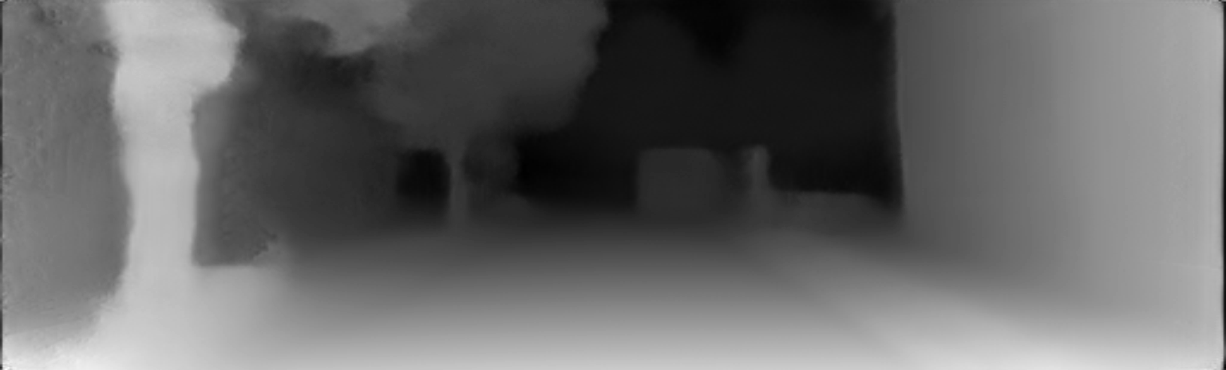} \includegraphics[width=0.24\linewidth]{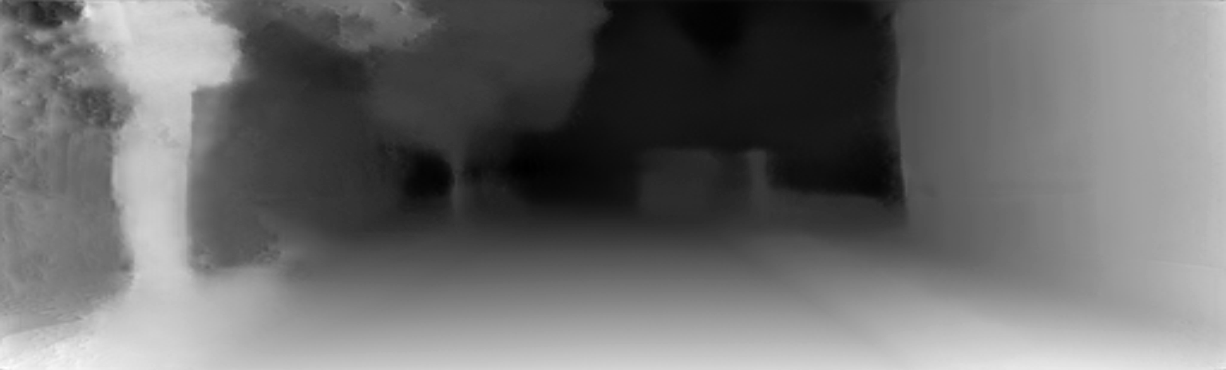}\\
\includegraphics[width=0.24\linewidth]{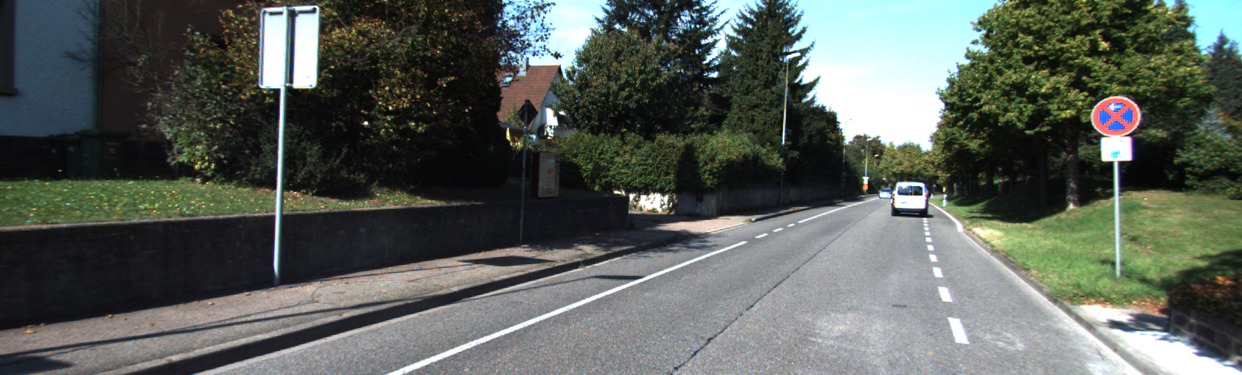} \includegraphics[width=0.24\linewidth]{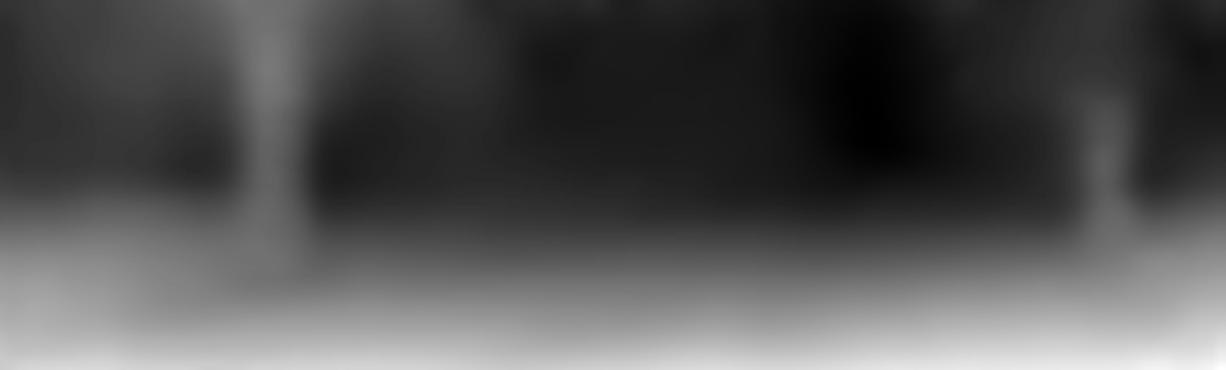} \includegraphics[width=0.24\linewidth]{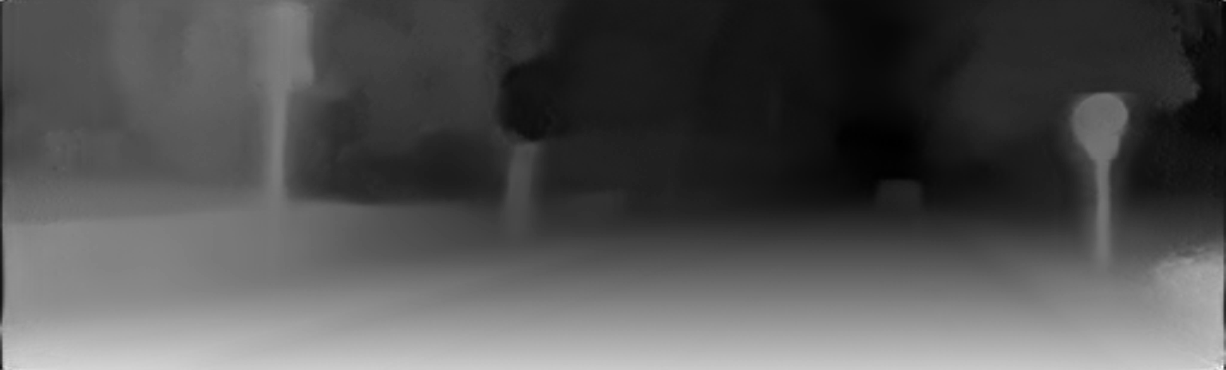} \includegraphics[width=0.24\linewidth]{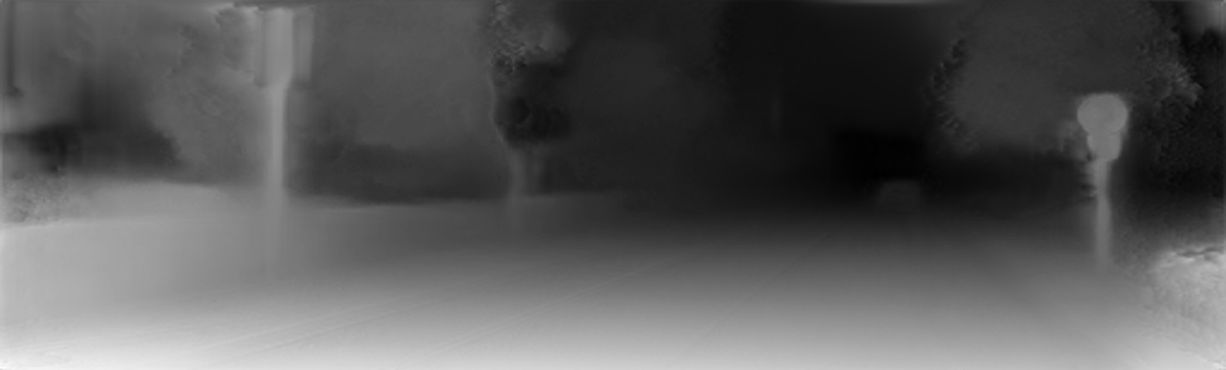}\\
\includegraphics[width=0.24\linewidth]{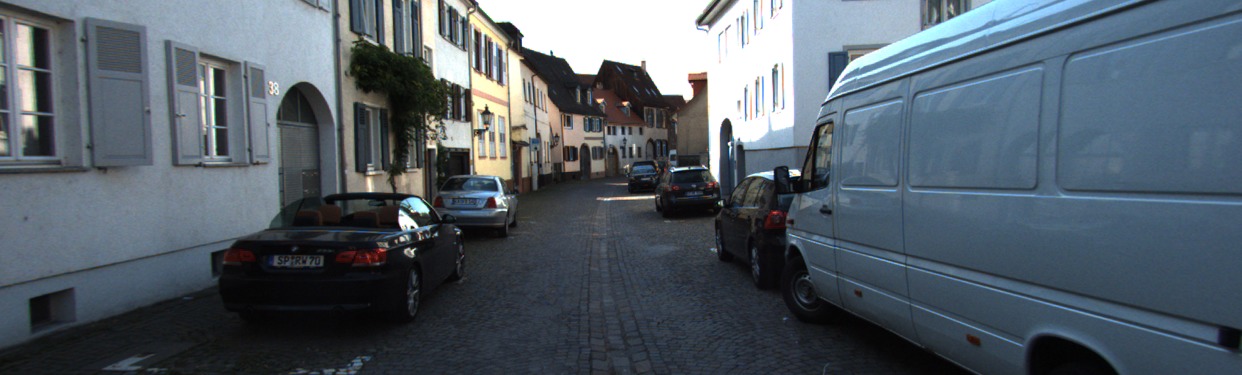} \includegraphics[width=0.24\linewidth]{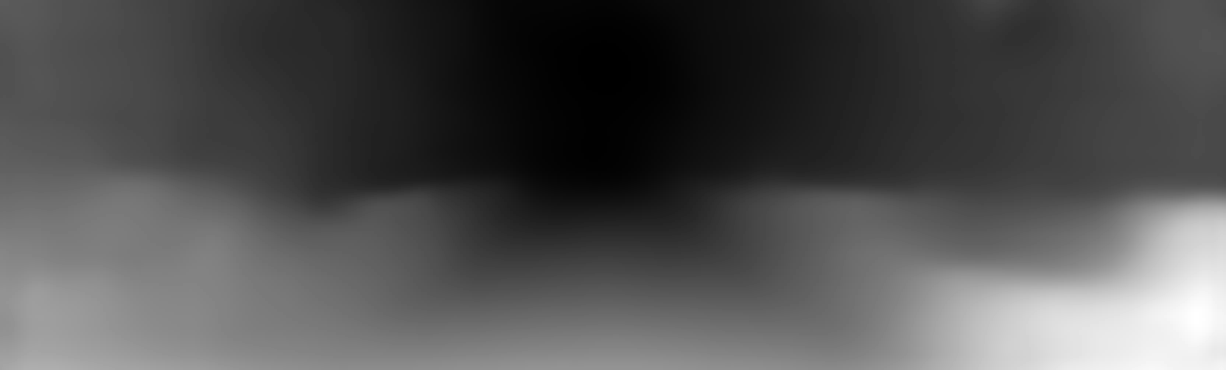} \includegraphics[width=0.24\linewidth]{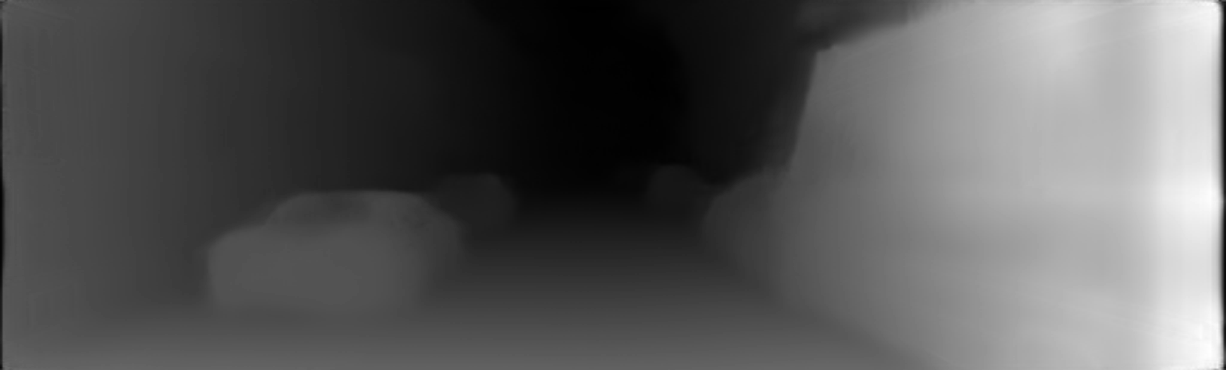} \includegraphics[width=0.24\linewidth]{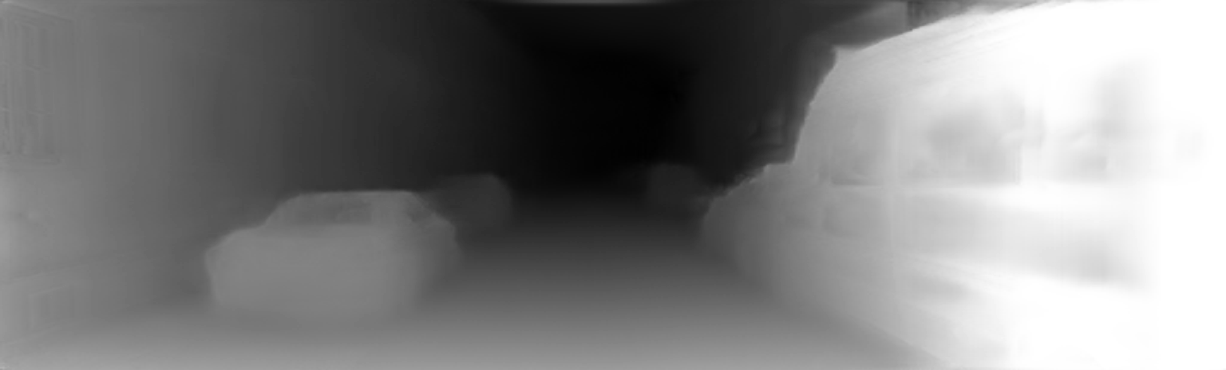}\\
\includegraphics[width=0.24\linewidth]{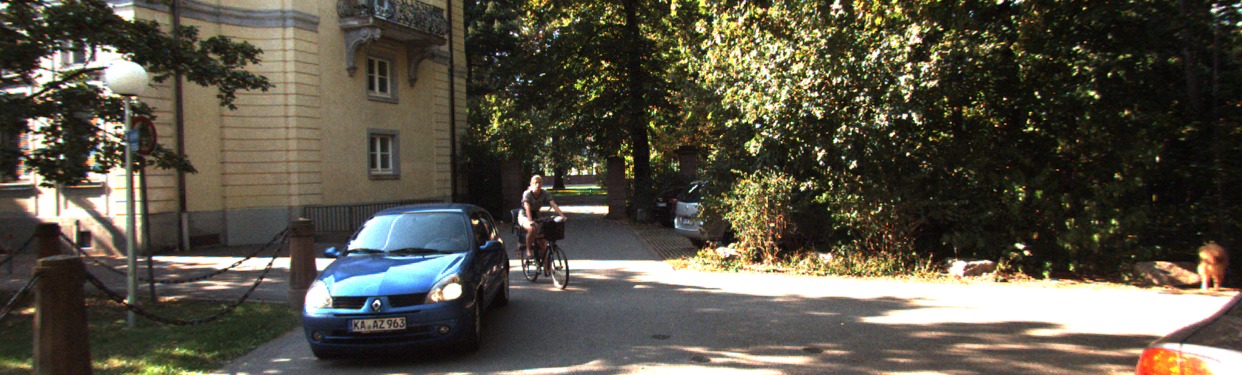} \includegraphics[width=0.24\linewidth]{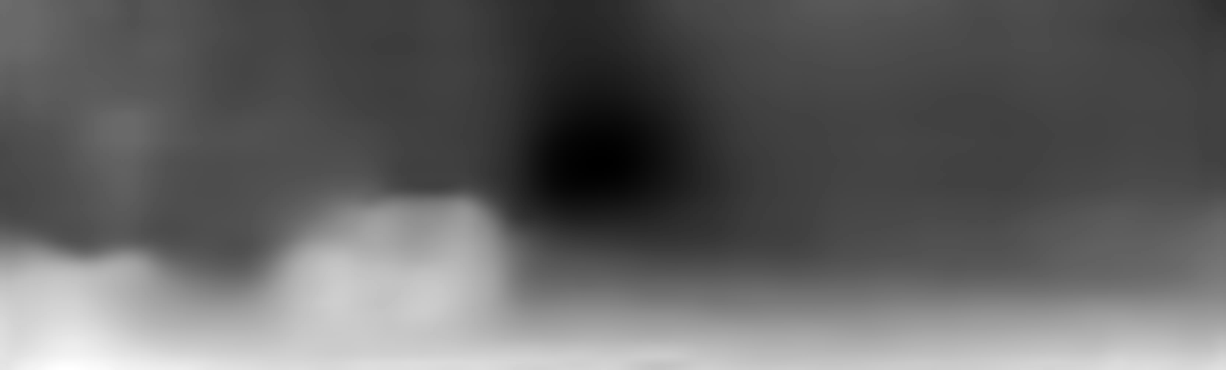} \includegraphics[width=0.24\linewidth]{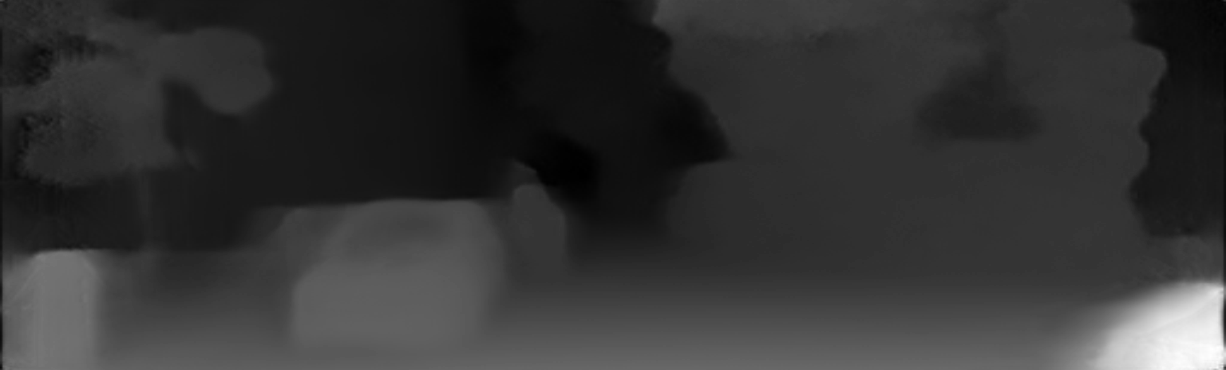} \includegraphics[width=0.24\linewidth]{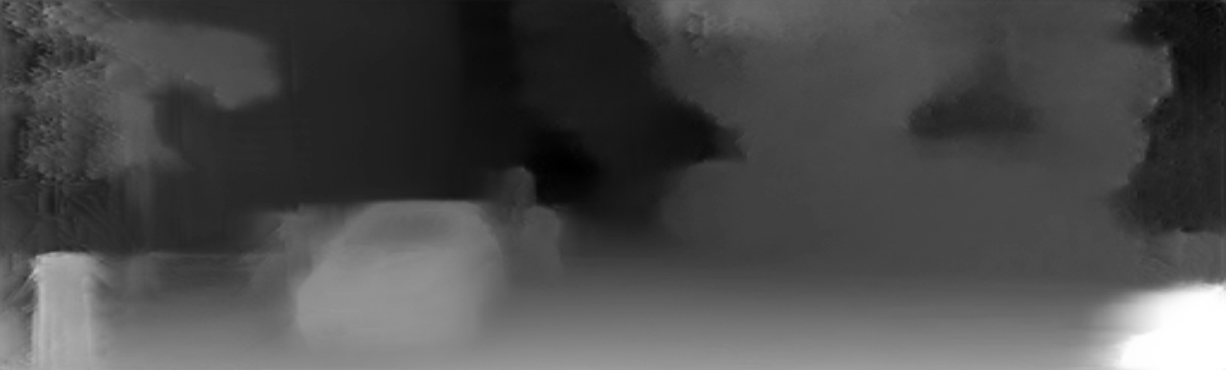}\\
\includegraphics[width=0.24\linewidth]{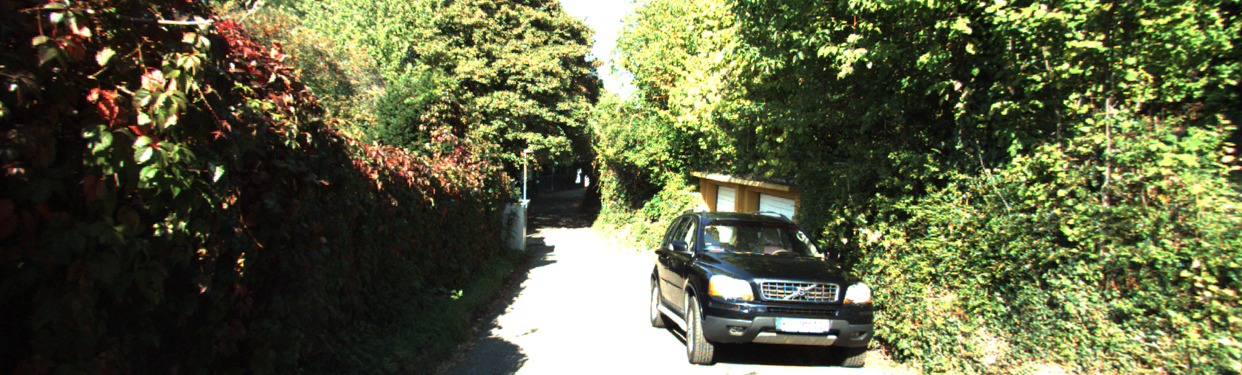} \includegraphics[width=0.24\linewidth]{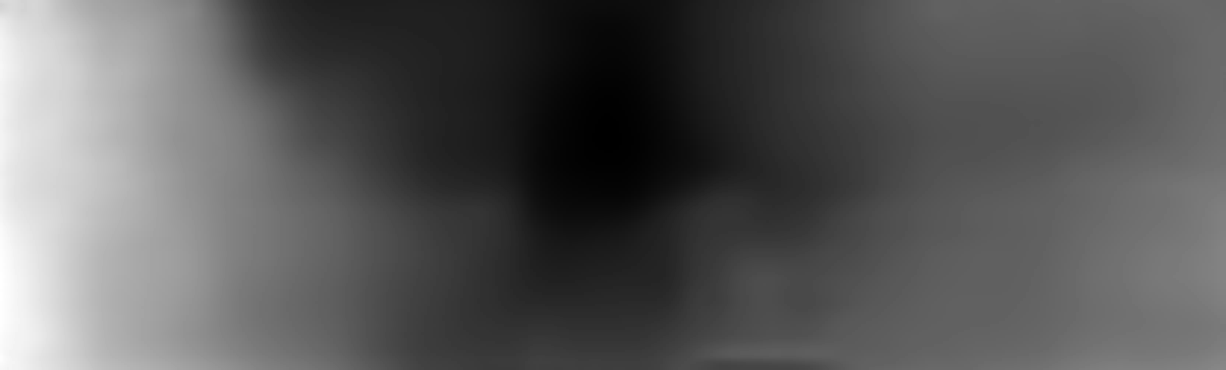} \includegraphics[width=0.24\linewidth]{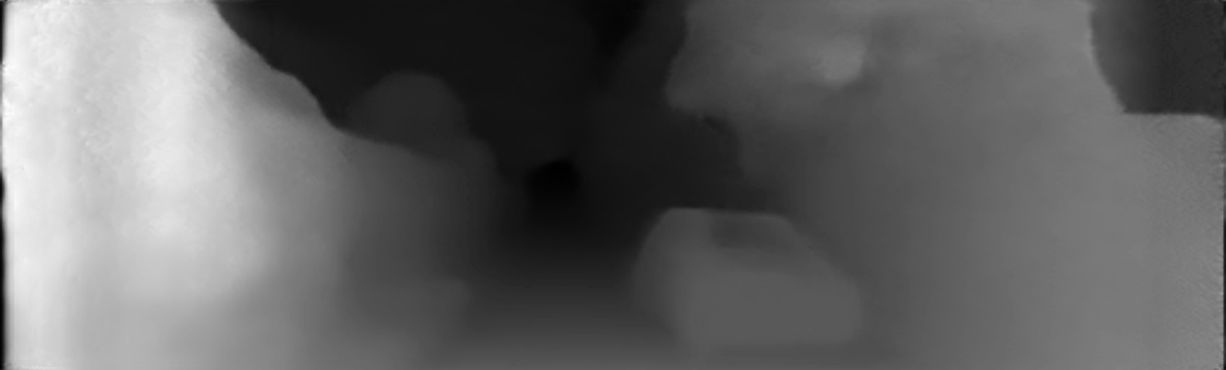} \includegraphics[width=0.24\linewidth]{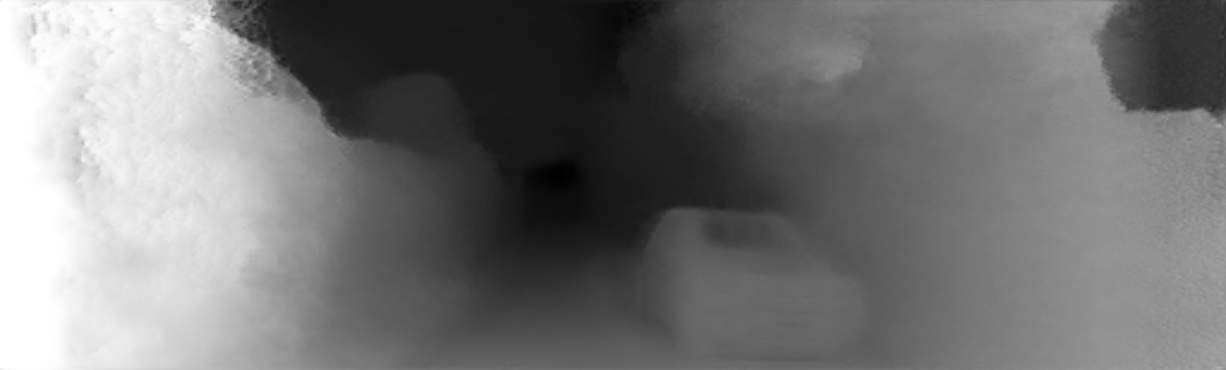}\\
\includegraphics[width=0.24\linewidth]{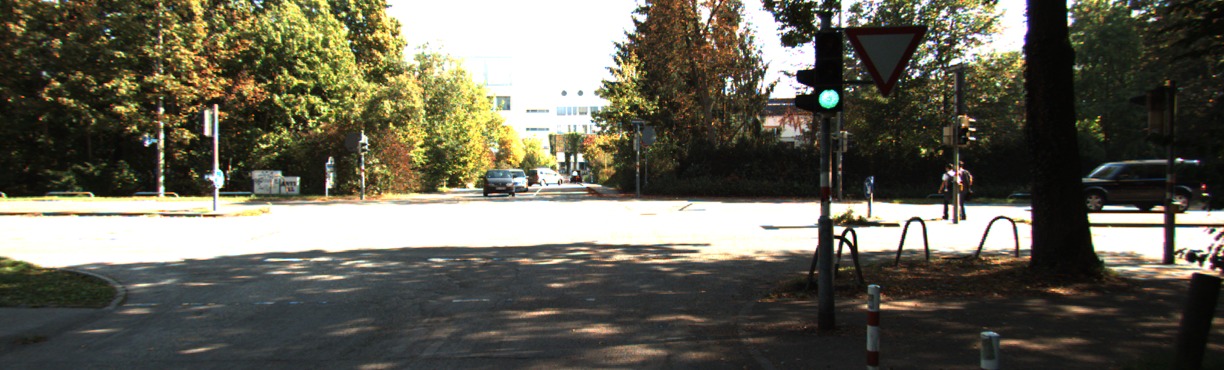} \includegraphics[width=0.24\linewidth]{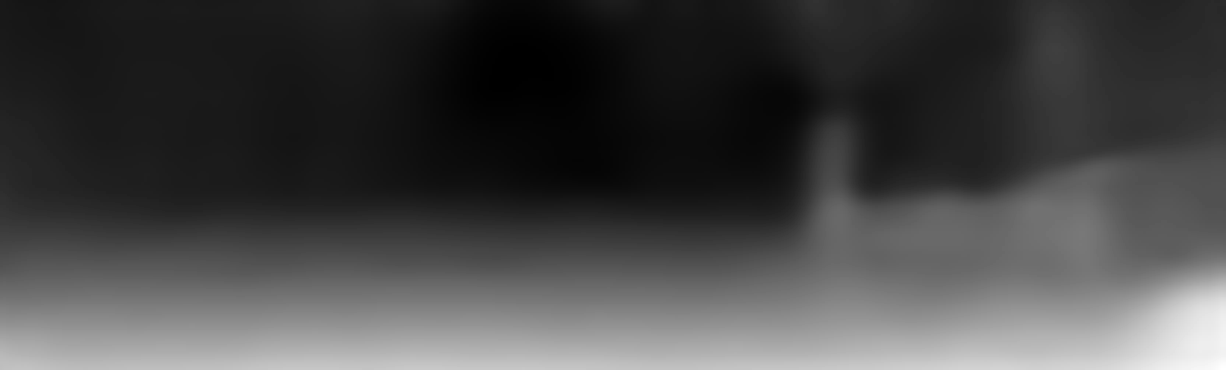} \includegraphics[width=0.24\linewidth]{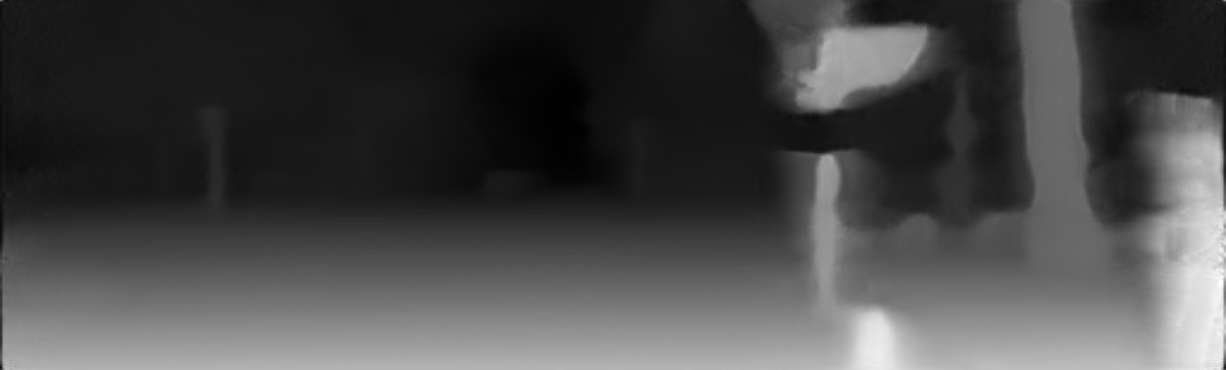} \includegraphics[width=0.24\linewidth]{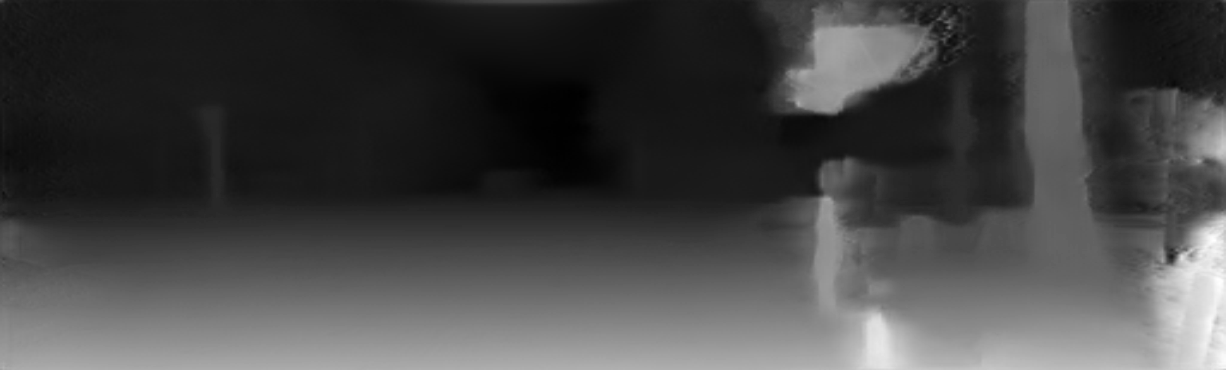}\\
\includegraphics[width=0.24\linewidth]{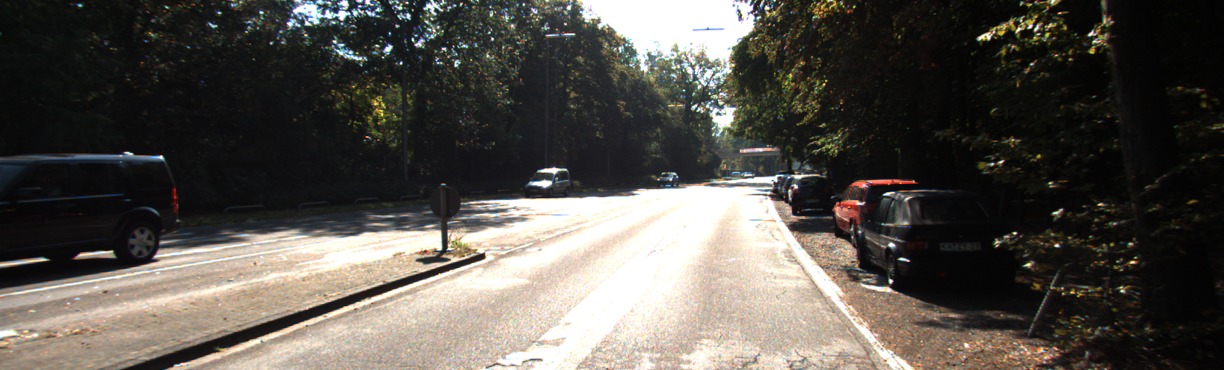} \includegraphics[width=0.24\linewidth]{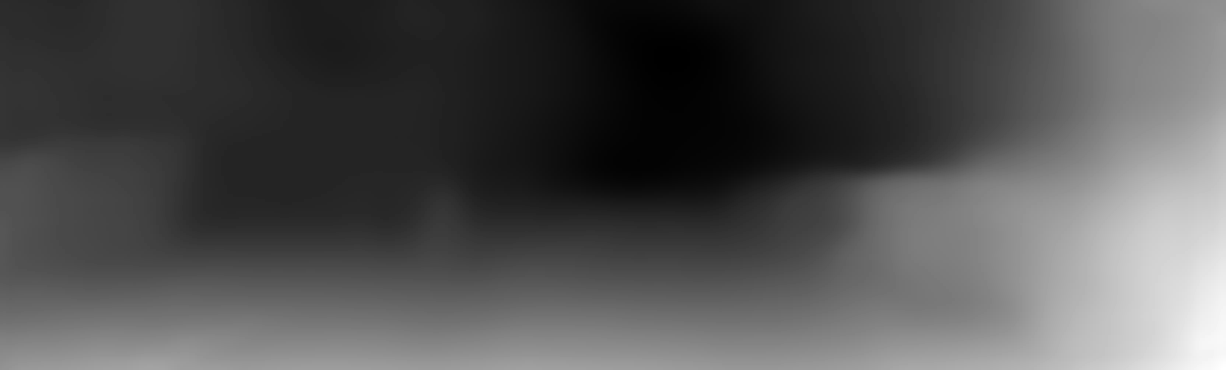} \includegraphics[width=0.24\linewidth]{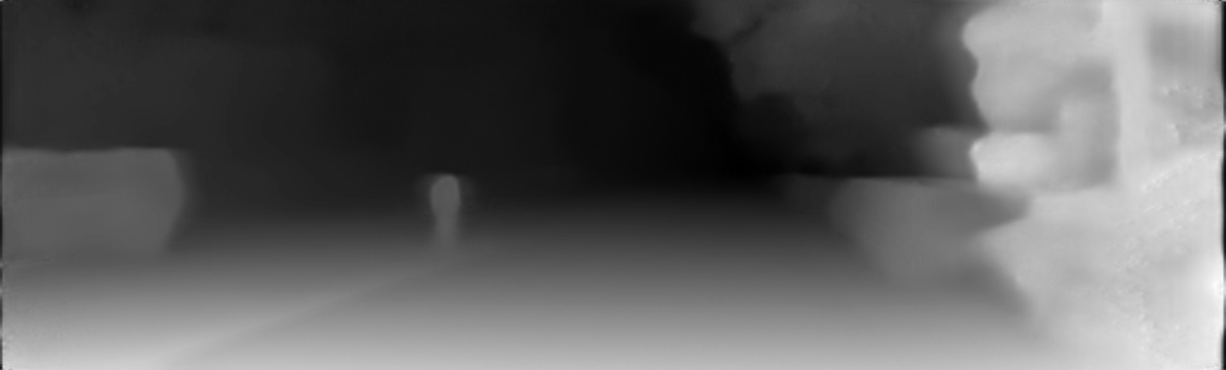} \includegraphics[width=0.24\linewidth]{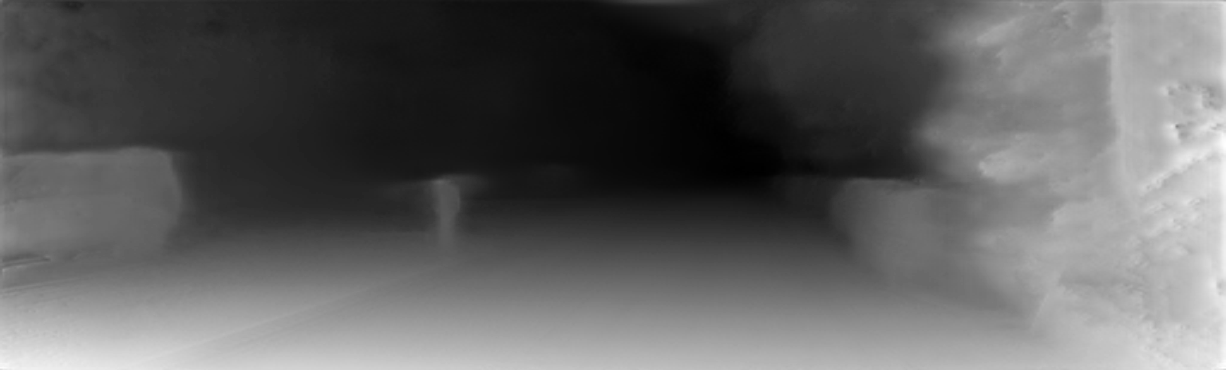}\\
\includegraphics[width=0.24\linewidth]{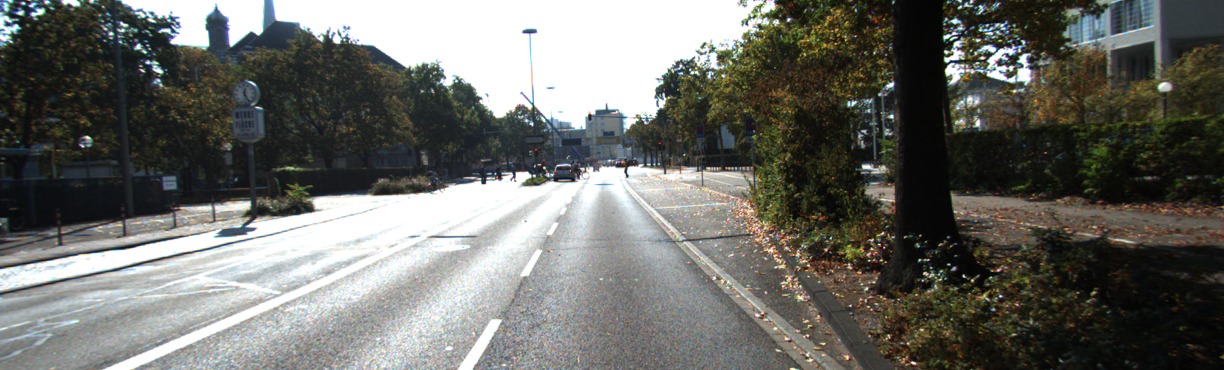} \includegraphics[width=0.24\linewidth]{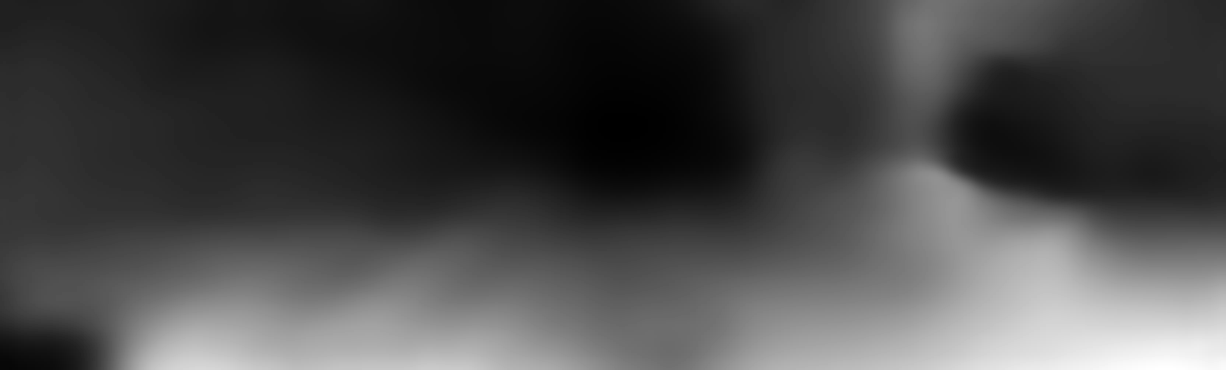} \includegraphics[width=0.24\linewidth]{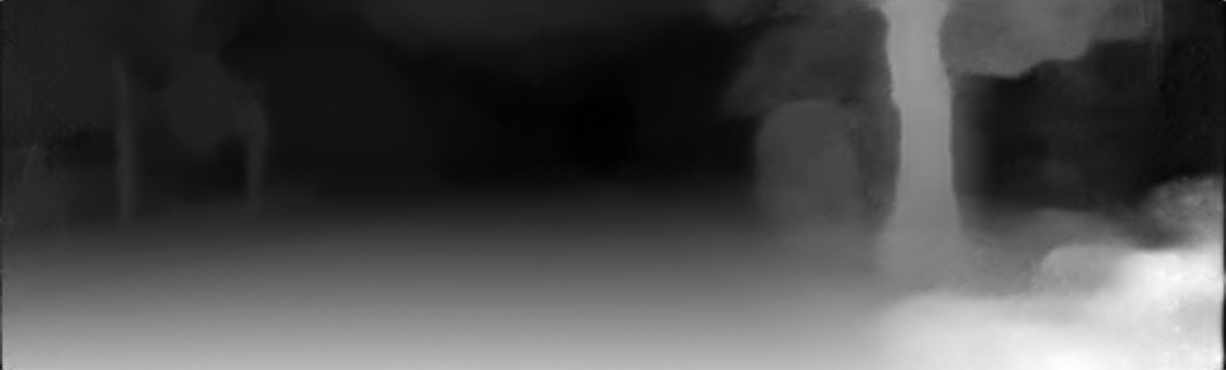} \includegraphics[width=0.24\linewidth]{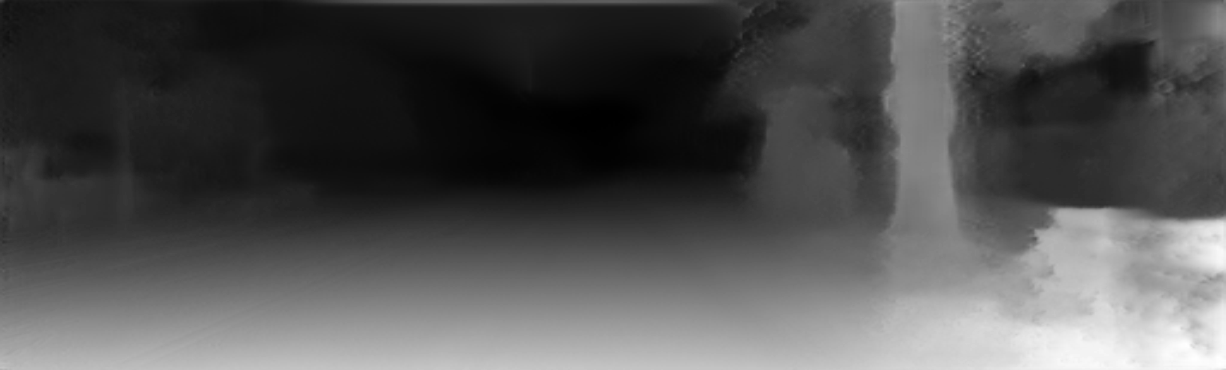}\\
\includegraphics[width=0.24\linewidth]{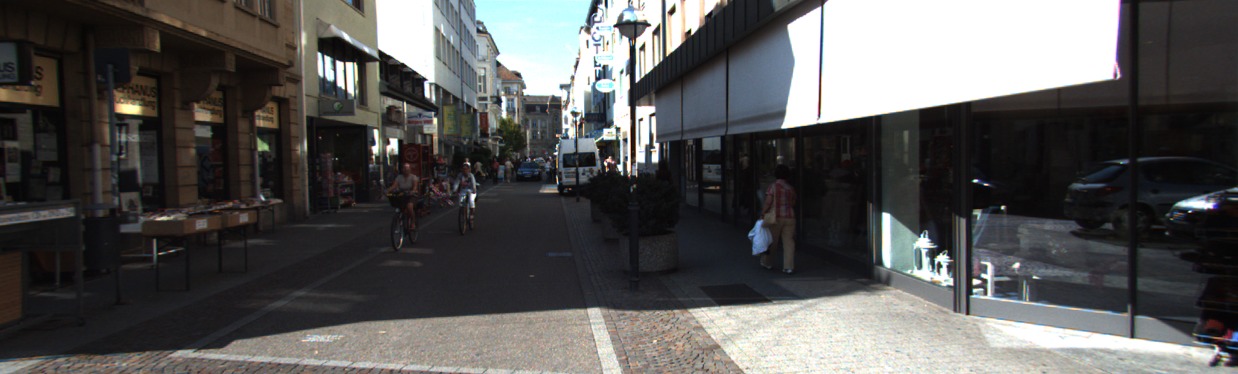} \includegraphics[width=0.24\linewidth]{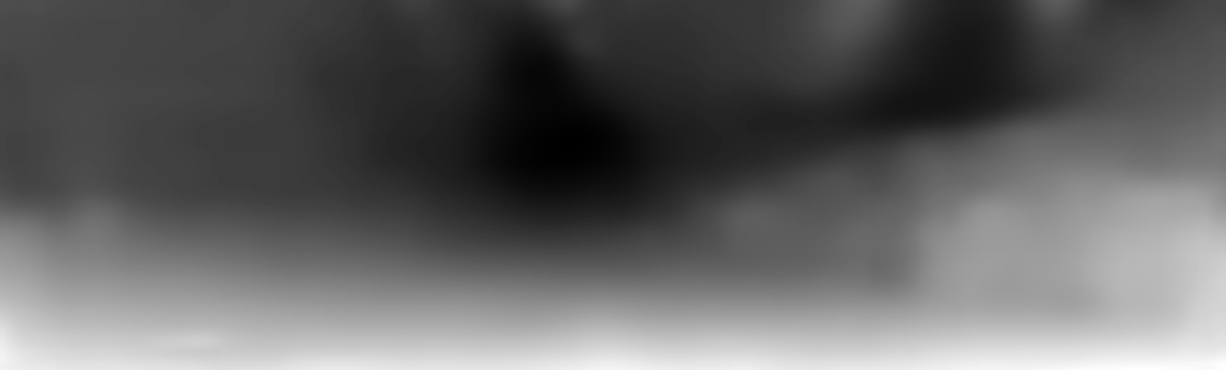} \includegraphics[width=0.24\linewidth]{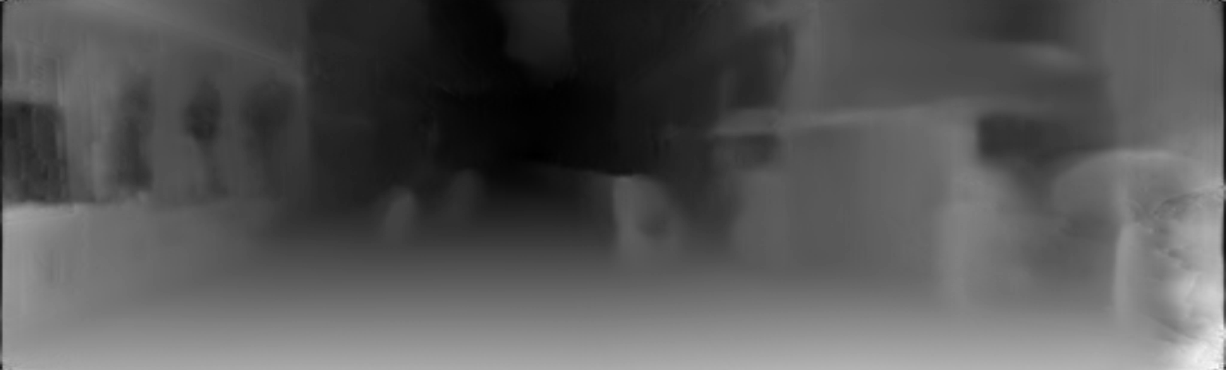} \includegraphics[width=0.24\linewidth]{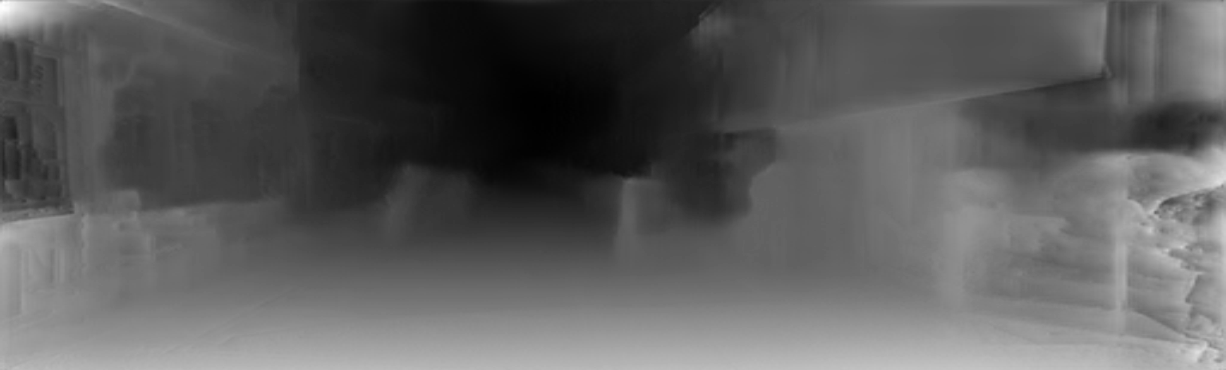}\\
\includegraphics[width=0.24\linewidth]{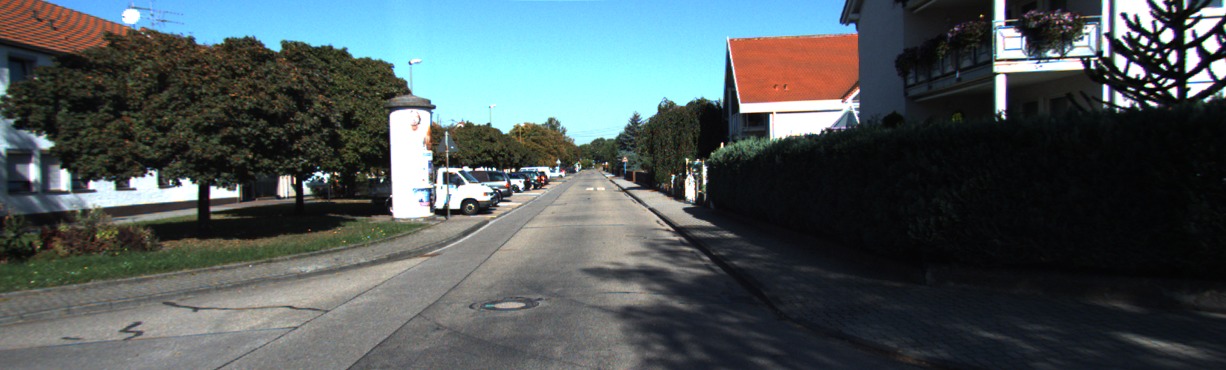} \includegraphics[width=0.24\linewidth]{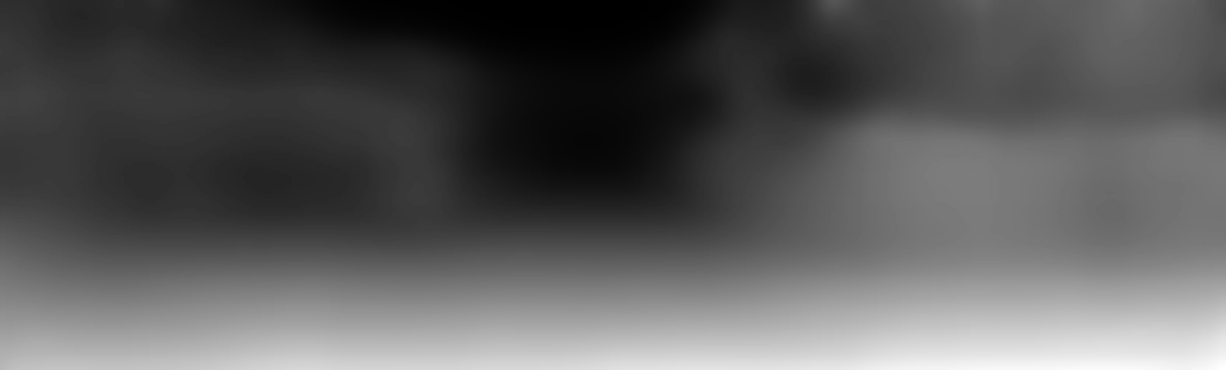} \includegraphics[width=0.24\linewidth]{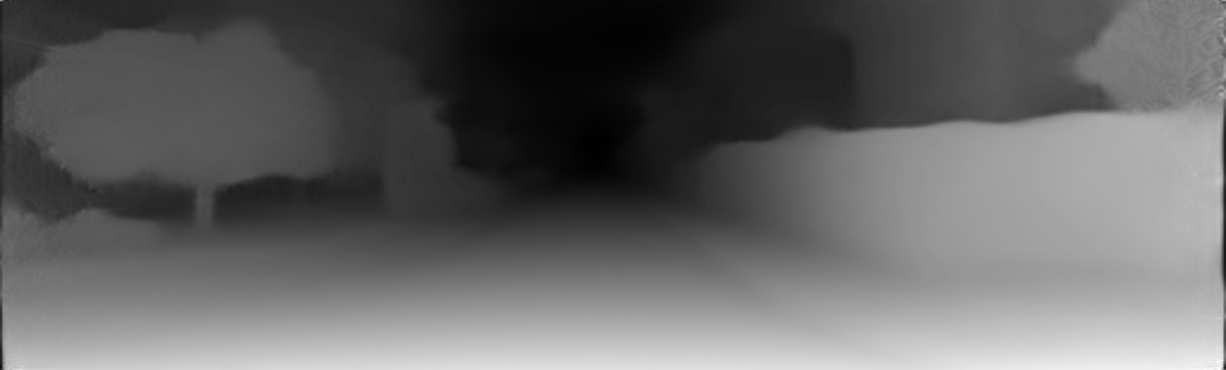} \includegraphics[width=0.24\linewidth]{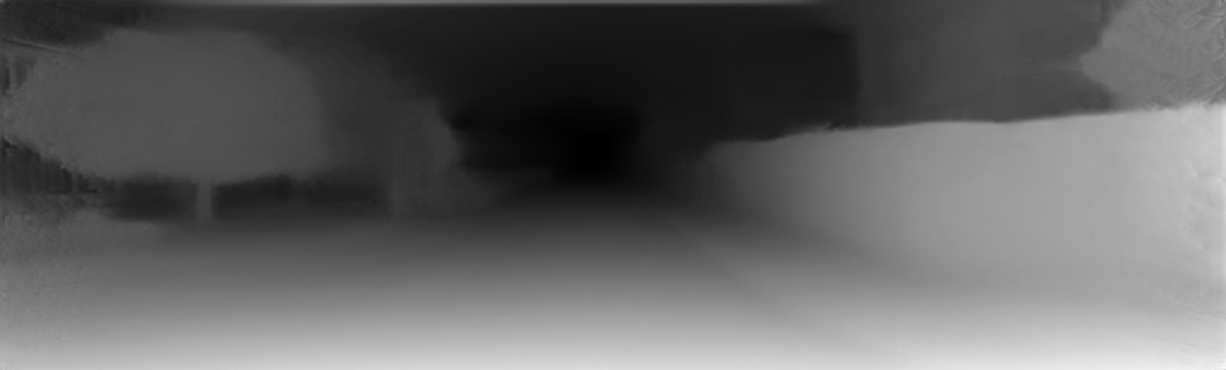}\\
\includegraphics[width=0.24\linewidth]{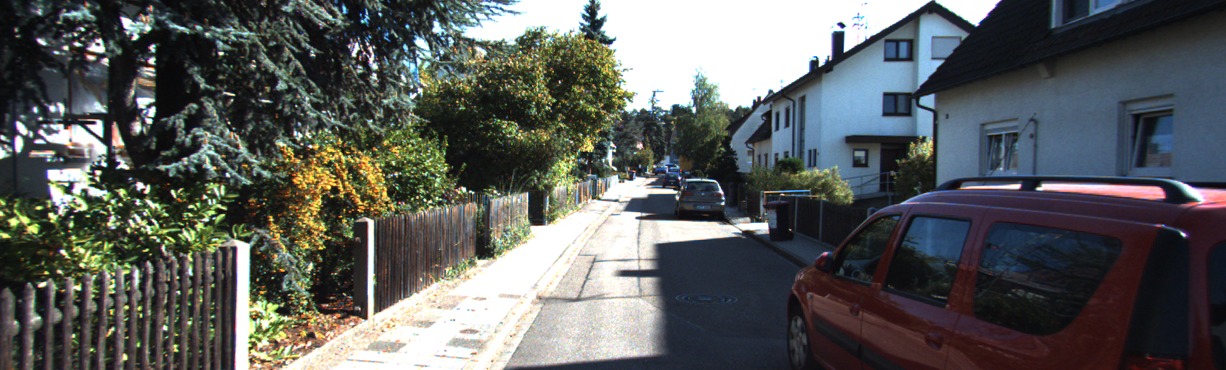} \includegraphics[width=0.24\linewidth]{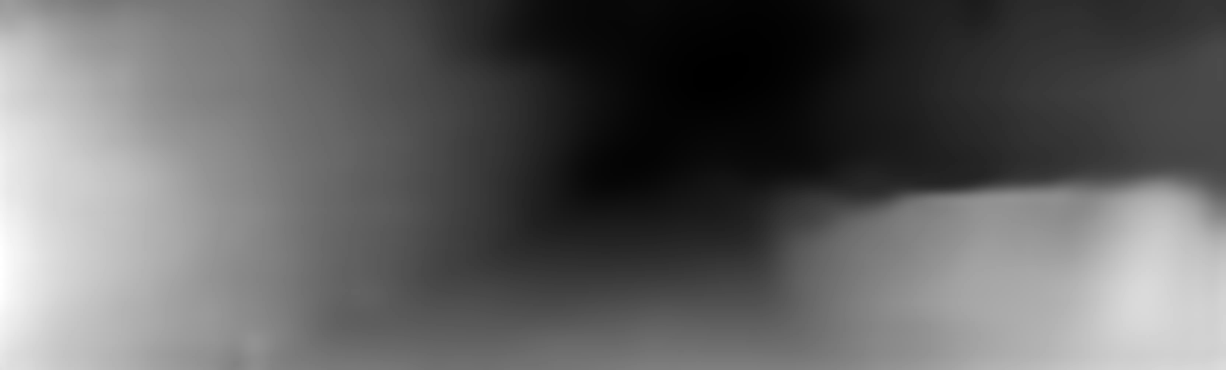} \includegraphics[width=0.24\linewidth]{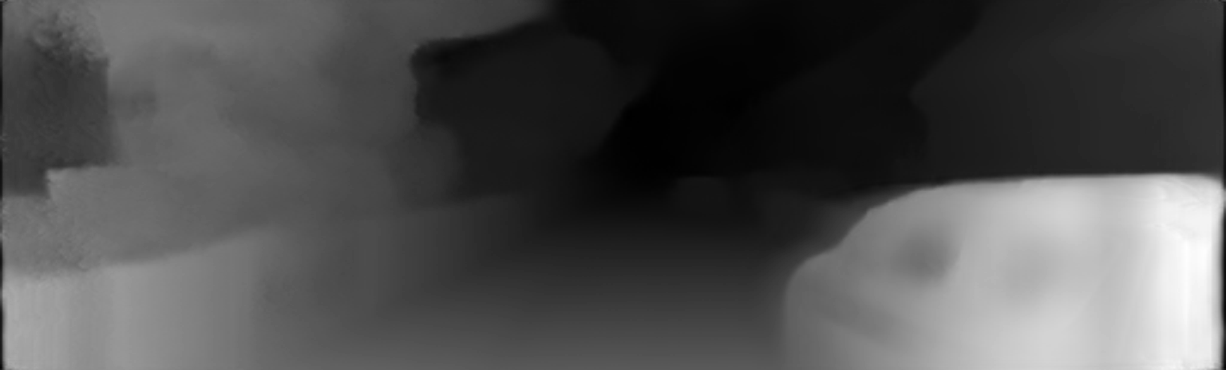} \includegraphics[width=0.24\linewidth]{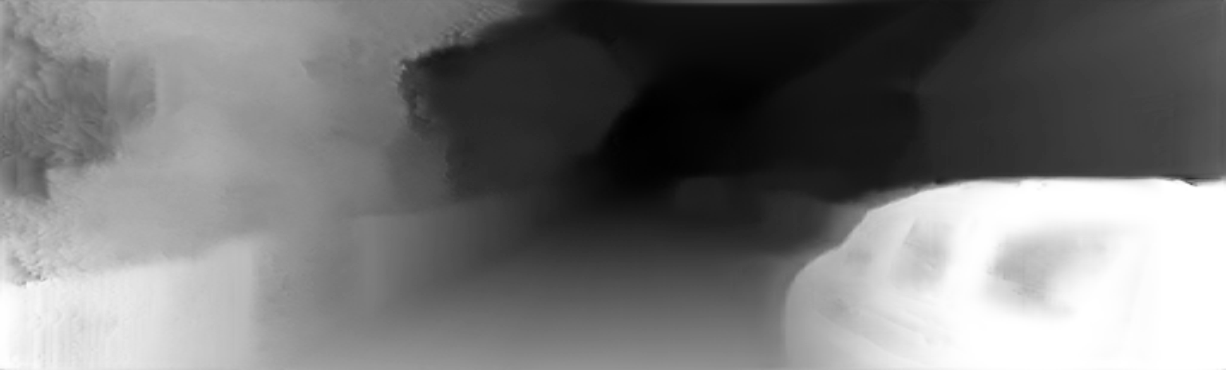}\\
\includegraphics[width=0.24\linewidth]{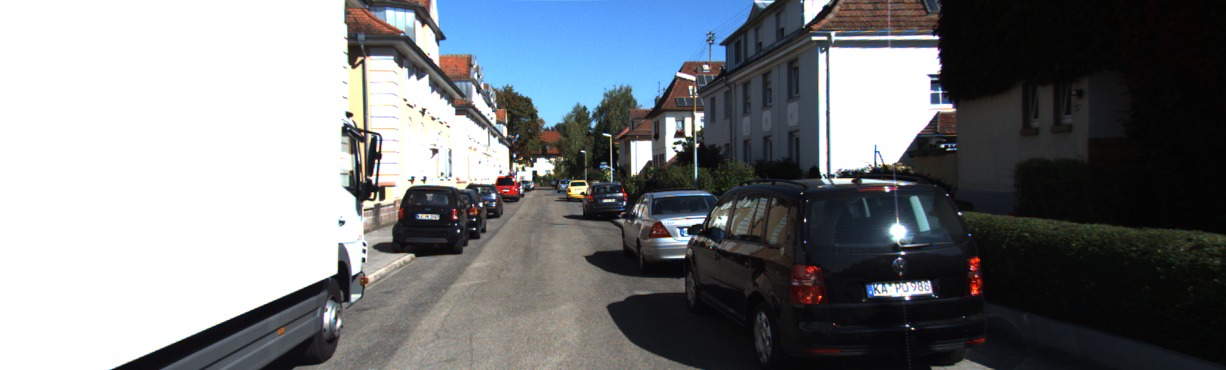} \includegraphics[width=0.24\linewidth]{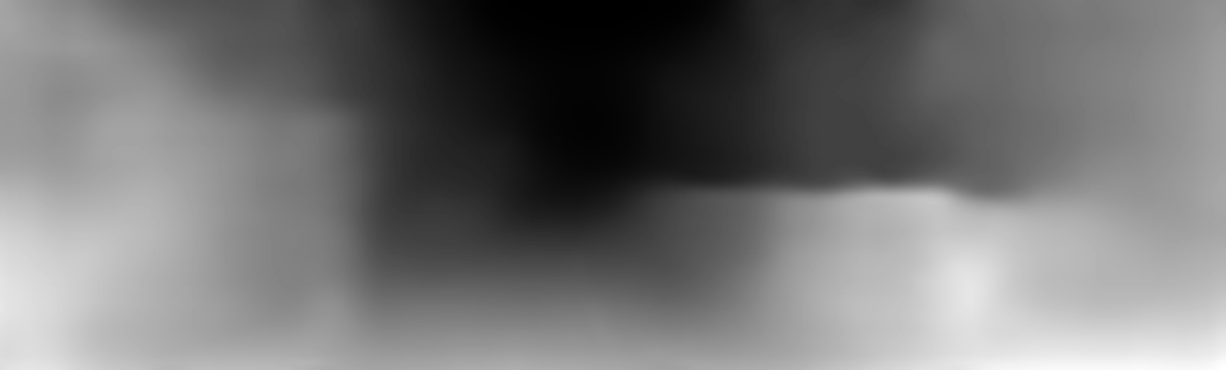} \includegraphics[width=0.24\linewidth]{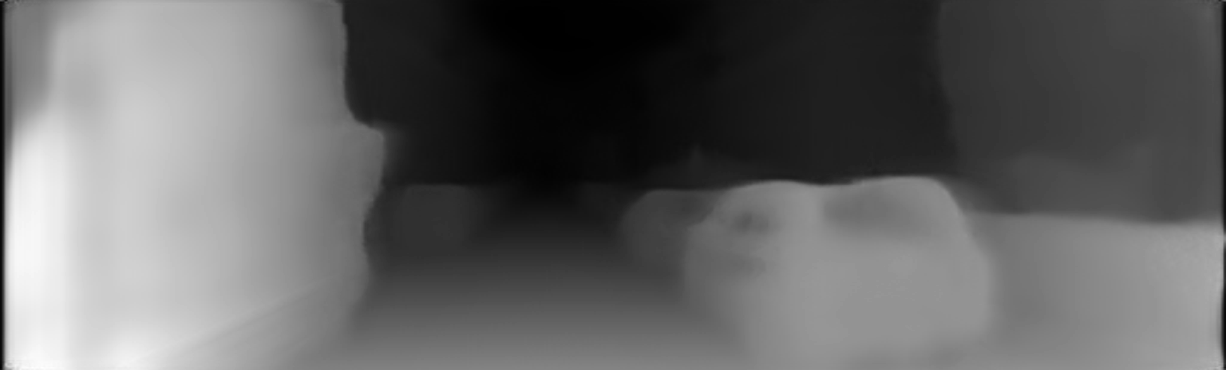} \includegraphics[width=0.24\linewidth]{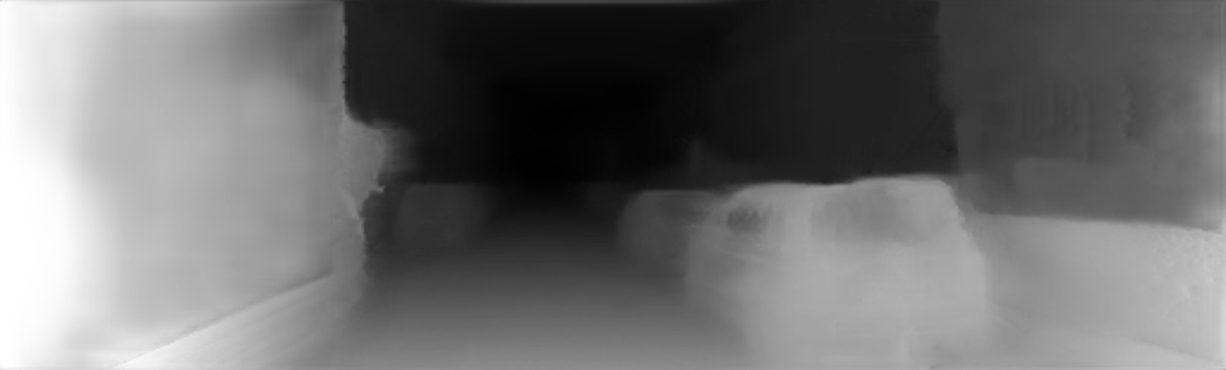}\\
\includegraphics[width=0.24\linewidth]{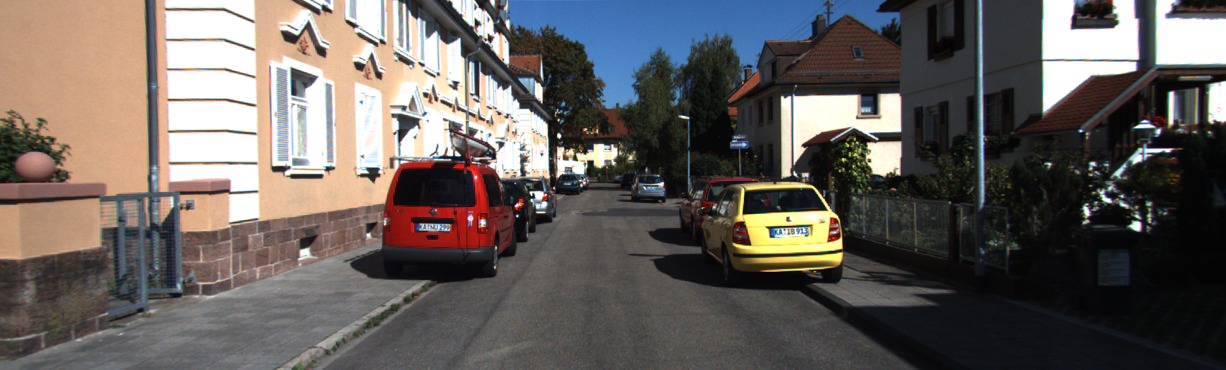} \includegraphics[width=0.24\linewidth]{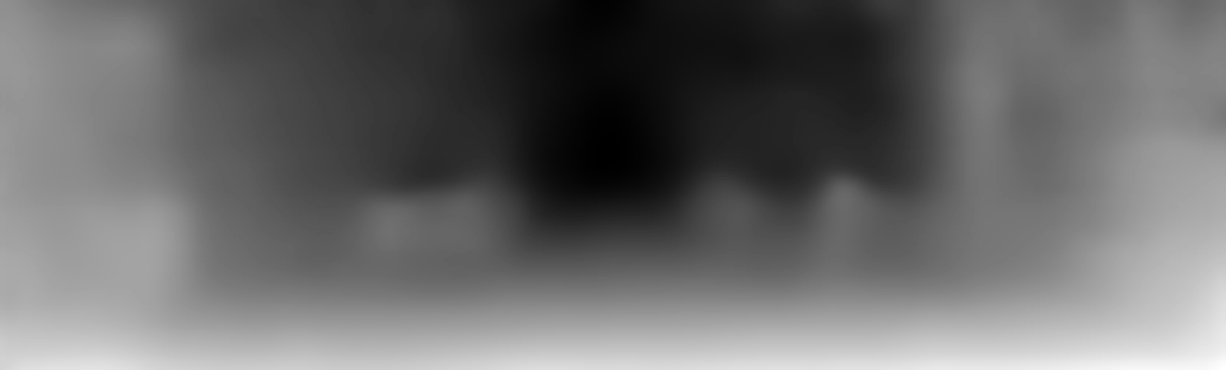} \includegraphics[width=0.24\linewidth]{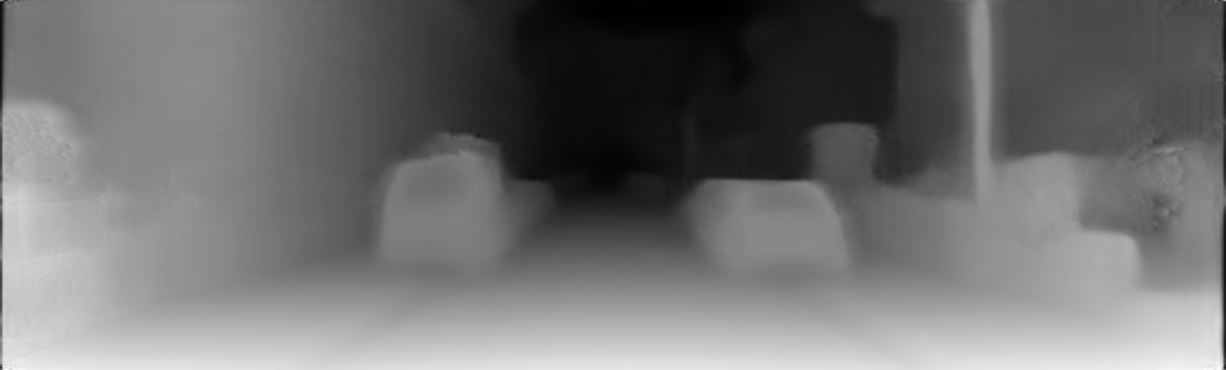} \includegraphics[width=0.24\linewidth]{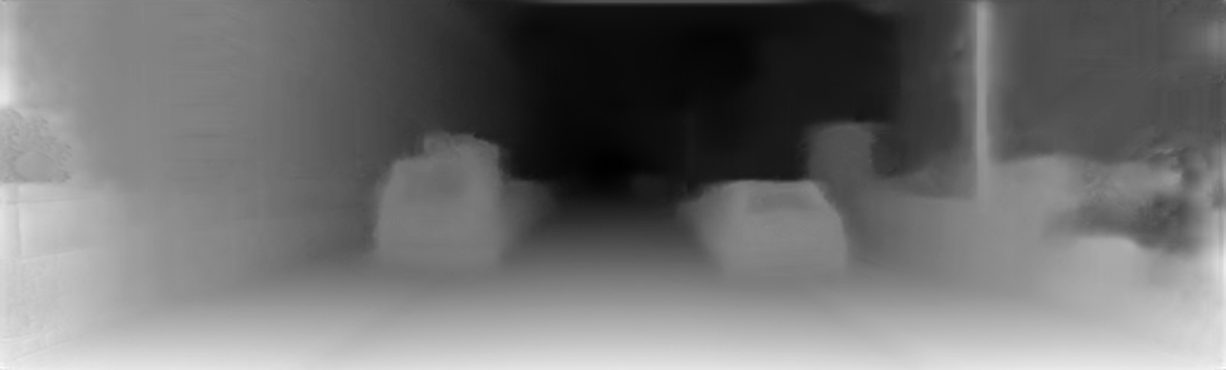}\\
\includegraphics[width=0.24\linewidth]{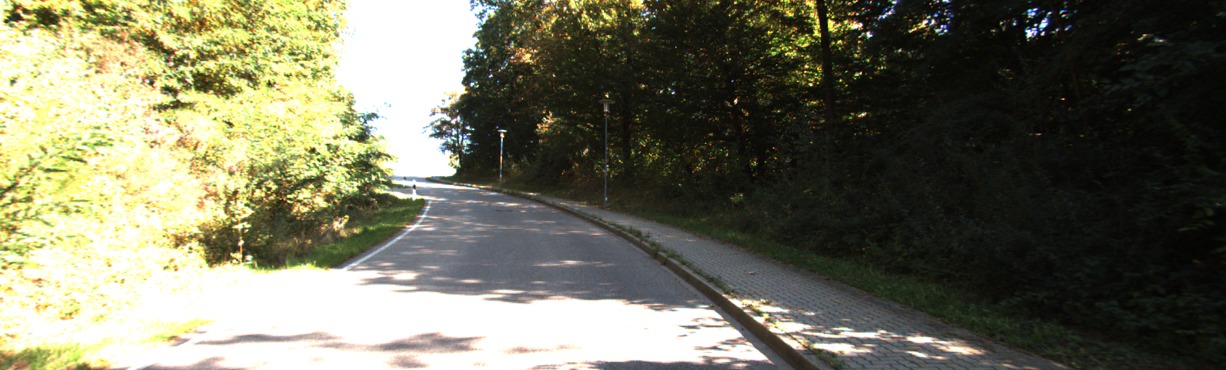} \includegraphics[width=0.24\linewidth]{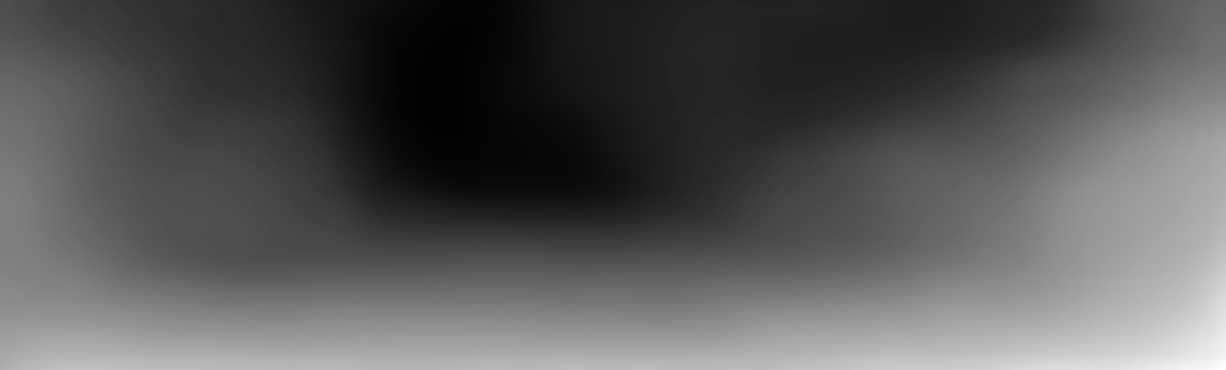} \includegraphics[width=0.24\linewidth]{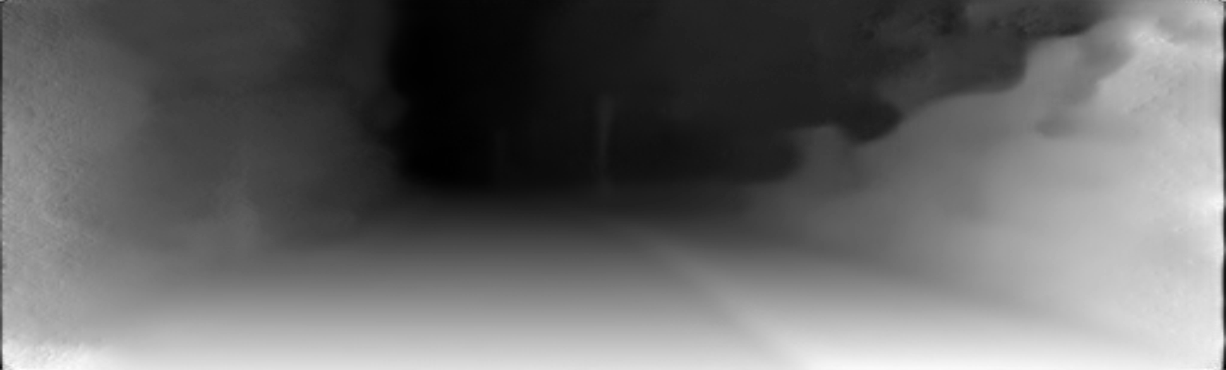} \includegraphics[width=0.24\linewidth]{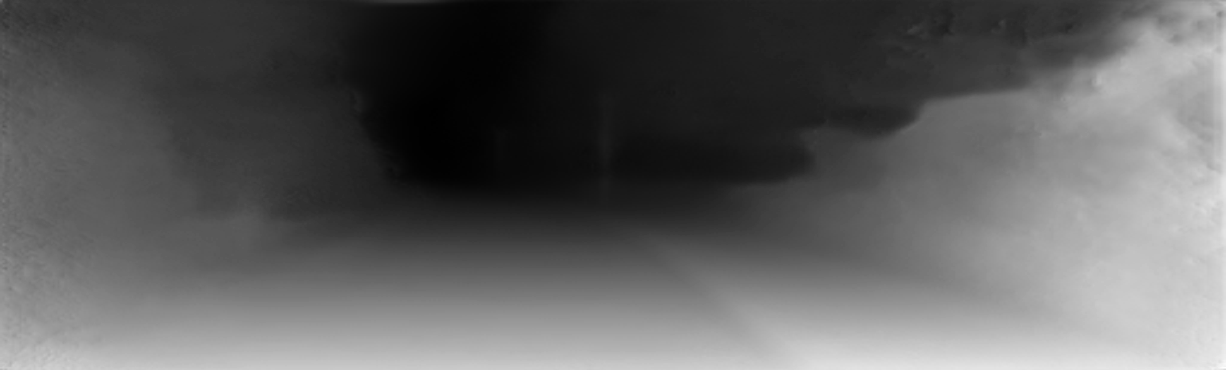}\\
\includegraphics[width=0.24\linewidth]{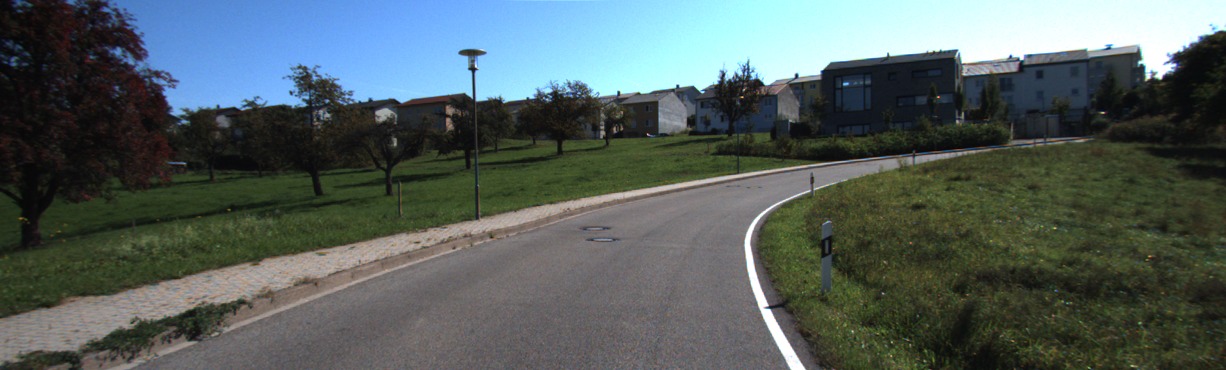} \includegraphics[width=0.24\linewidth]{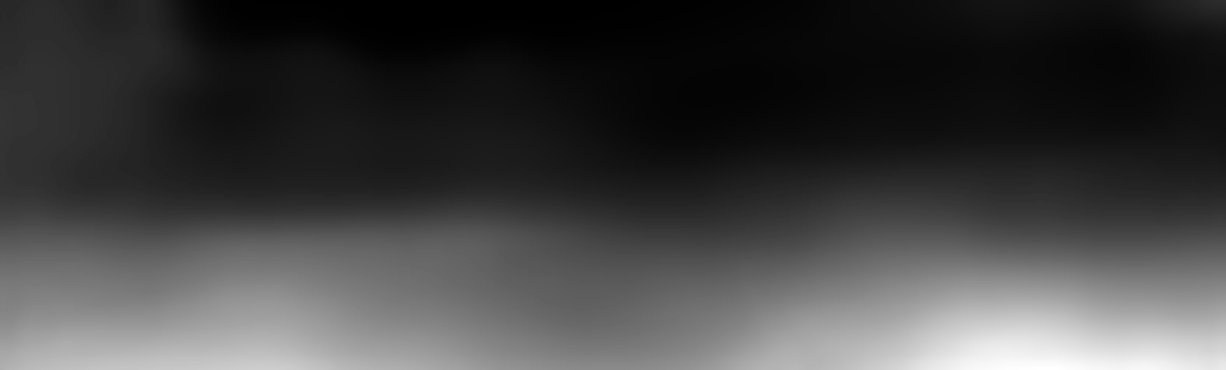} \includegraphics[width=0.24\linewidth]{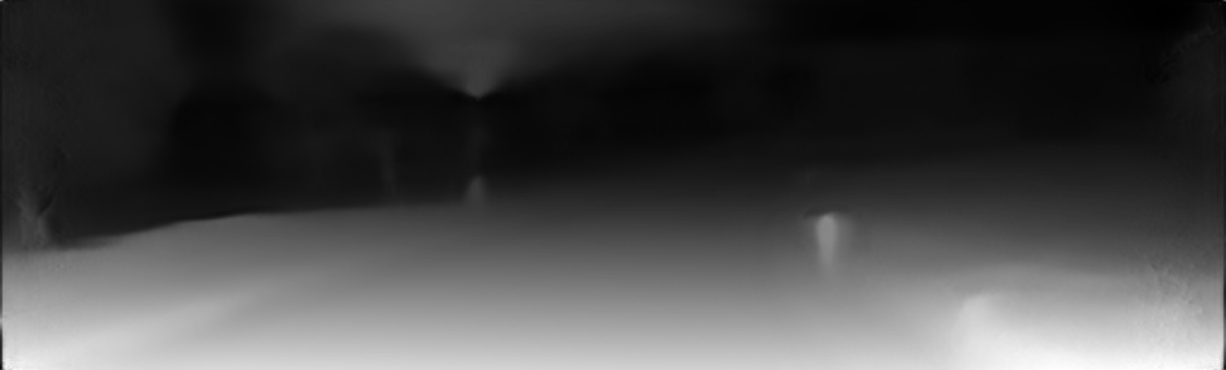} \includegraphics[width=0.24\linewidth]{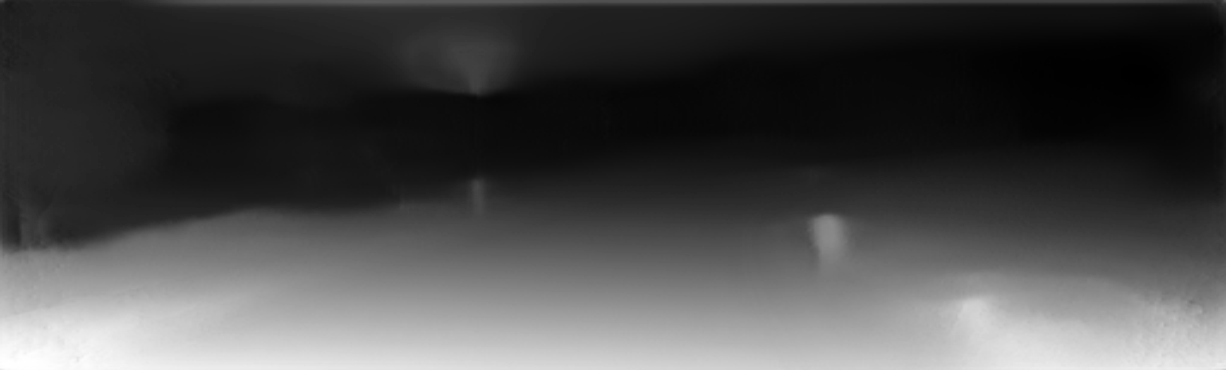}
\caption{Shown are 20 randomly chosen images from the dataset visualizing the results of different models. The output of Godards model is quite similar to ours due to overlapping loss functions being used. Note that for this visualisation our depth maps have been inverted.}
\label{fig:comparison}
\end{figure}

To compare this model to different approaches two splits of the KITTI 2015 dataset \cite{geiger_are_2012} are used due to its popularity. Qualitative comparisons are shown in figure \ref{fig:comparison}. The depth map used for evaluation is the one generated from the right image, the reasoning can be found in section \ref{par:sampler}.

\subsection{Model Details}
The network is implemented in Tensorflow \cite{abadi_tensorflow:_2016}. Training on the Kitti data-set, consisting of about 30.000 images, for 50 epochs takes close to 50 hours on a Titan X GPU, inference on a GTX 970 achieves a speed of roughly 19 frames per second for images with a size of 512$\times$ 256. \\
The output disparities are given by a sigmoid layer scaled with $d_{max}=0.3$. For the activation function, the network uses exponential liner units \cite{clevert_fast_2015} for the part estimating the depth and RELUs \cite{nair_rectified_2018} for pose estimation. For upscaling a nearest neighbor upsampling followed by a convolution is used. The model is trained from scratch with a batchsize of 8 using Adam \cite{kingma_adam:_2014} with $\beta_1=0.9$, $\beta_2=0.999$ and $\epsilon=10^{-8}$. The initial learning rate is kept at $\lambda=10^{-4}$ for 30 Epochs and then halved every 10 Epochs. \\
Data augmentation is performed on half of the images by randomly flipping the image horizontally and/or changing gamma, brightness and shifting color in the interval from [0.8,1.2], [0.5,2] and [0.8,1.2] respectively.

\subsection{Post-Processing}
We asume the current pose of the left camera to be incident with the world system. This causes stereo occlusion in the left part of the disparity map, since after warping there is no information to work with, as the objects are not in the image anymore. Therefore during test time an additional inference is run on the horizontally flipped image. Flipping this disparity map back lines it up with the original disparity map. In the flipped map the missing part caused by stereo occlusion is located on the right side instead. 
In a post processing step the final result is composed by a ramped vertical stripe (5\% of the image size) from the infered disparity map of the flipped image, the average of both maps and a sencond ramped stripe (also 5\%) from the unflipped image.

\subsection{Performance}
The model is compared to other models on two splits of the KITTI dataset. The differences are shown in table \ref{tab:splits}. The evaluation of both splits are shown in tables \ref{tab:Kitti_split} and \ref{tab:Eigen_split}. \\
Our model achieves a superior performance in 6 out of 8 categories in the split provided by Kitti, but performs poorly on the absolute error and the D1-all metric. This is due to a less smooth output of our model and the evaluation in disparity space requiring an inverse transformation which causes big errors at lower depth. \\
On Eigen's split we achieve superior performance in all metrics.

\begin{table*}[t]
\begin{center}

\begin{tabular}{|l|l|l|}
\hline
 & KITTI Split \cite{geiger_are_2012} & Eigen Split \cite{eigen_depth_2014} \\ \hline
Number of images & 200 & 697 \\
Ground truth & Velodyne points + 3D car models & Velodyne points \\
Evaluates & Disparity & Depth \\
Metric & Eigen \cite{eigen_depth_2014} + D1-all from KITTI \cite{geiger_are_2012} & Eigen \cite{eigen_depth_2014} \\
Maximum depth & 80m & 50m \& 80m \\ 
Inaccuracies & At CAD model boundaries & \makecell[l]{At occlusions between \\ the LIDAR and the camera}
\\ \hline
\end{tabular}
\caption{The table compares the splits of KITTI 2015 and Eigen et al. }
\label{tab:splits}
\end{center}
\end{table*}

\begin{table*}[t]
\begin{center}
\begin{tabular}{|l|l|l|l|l|l|l|l|l|}
\hline
Method & Abs Rel & Sq RMS & RMSE & RMSE log & D1-all & $\delta <1.25$ & $\delta<1.25^2$ &$\delta <1.25^3 $\\ \hline
Godard  \cite{godard_unsupervised_2016} & \textbf{0.124} & 1.388 & 6.125 & 0.217 & \textbf{30.272} & 0.841 & 0.936 & 0.975 \\
Ours & 0.134 & \textbf{1.195} & \textbf{5.732} & \textbf{0.208} & 39.628 & \textbf{0.843} & \textbf{0.945} & \textbf{0.980}  \\ \hline
\end{tabular}
\caption{The table compares the results of Godard's model, taken from their paper and ours on the 200 validation images from the KITTI 2015 set. Both models are trained on the KITTI dataset. }
\label{tab:Kitti_split}
\end{center}
\end{table*}

\begin{table*}[t]
\begin{center}
\begin{tabular}{|l|l|l|l|l|l|l|l|}
\hline
Method & Abs Rel & Sq Rel & RMSE & RMSE log & $\delta <1.25$ & $\delta < 1.25^2$ &$\delta <1.25^3 $\\ \hline
Godard  \cite{godard_unsupervised_2016}   & 0.148 & 1.344 & 5.927 & 0.247  &0.803  & 0.922 & 0.964 \\
Eigen Fine  \cite{eigen_depth_2014}                &0.203 & 1.584 & 6.307 & 0.282  &0.702  & 0.890 & 0.958 \\
Zhou  \cite{zhou_unsupervised_2017}             &0.208 & 1.768 & 6.856 & 0.283 & 0.678  & 0.885 & 0.957 \\
Ours & \textbf{0.137} & \textbf{1.131} & \textbf{5.169} & \textbf{0.221}  & \textbf{0.839} & \textbf{0.939} & \textbf{0.972}  \\ \hline
Garg  \cite{garg_unsupervised_2016}              &0.169 & 1.080 & 5.104 & 0.273 & 0.740  & 0.904 & 0.962 \\
Ours capped & \textbf{0.131} & \textbf{0.845} & \textbf{3.973} & \textbf{0.209}  & \textbf{0.852} & \textbf{0.947} & \textbf{0.976}  \\ \hline
\end{tabular}
\caption{The table compares the results on the KITTI dataset using the Split of Eigen et al. \cite{eigen_depth_2014}. Numbers are taken from the respective publications. Note that the results of \cite{garg_unsupervised_2016} are capped at 50m. A comparison of our model capped to 50m can be found in the lower part of the table.}
\label{tab:Eigen_split}
\end{center}
\end{table*}

\newpage
\subsection{Qualitative Evaluation}
\label{sec:evaluation}
\begin{figure} 
\includegraphics[width=0.9\linewidth]{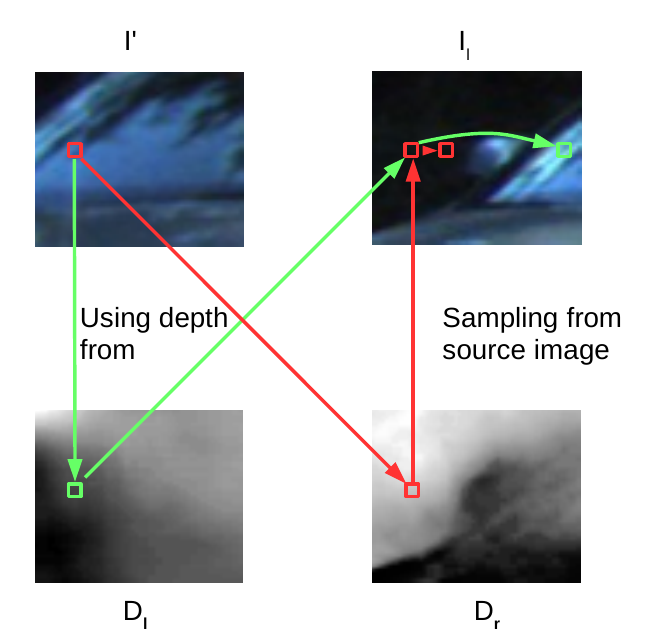}
\caption{The reconstruction using the inflated depth $D_l$ is shown in green, the reconstruction using $D_r$ is shown in red.}
\label{fig:inflation}
\end{figure}

\paragraph{Left Disparity Image}
\label{par:sampler}
The depth map corresponding to $I_l$ tends to inflate objects close to the camera. 
As the interpolator interpolates from surrounding pixel values based on the estimated depth at a point instead of moving points based on their depth, a point in the background uses nearby background pixels to interpolate even if it would actually be occluded. An example of this process is illustrated in figure \ref{fig:inflation}. Using the depth from $D_r$ causes only a small shift to find the pixels to interpolate from which is still on the background. With the inflated depth from $D_l$ the pixels to interpolate from are further away on the windshield. Note that the model is trained on image reconstruction as opposed to ground truth depth. Therefore the model favors an inflated depth map for a better reconstruction over a better depth map.
\paragraph{Explainability Mask}
Reconstructing slim objects like posts is a hard task as the depth prediction has to be very accurate and is prone to errors. The explainability mask, while intended for moving object and occlusions, has a  loss function scaling linearly with the pixel count it covers. Ultimately the error from covering up small and slim objects with the explainability masks results in lower errors than trying to reconstruct it.

\section{Conclusion}
We combined structure from motion and structure from stereo in an unsupervised learning approach. The model is split into a disparity estimation and pose estimation module using both to warp images to their stereo counterparts and future frames. The warped images are compared to their correspondent real images with SSIM and l1 and the resulting disparity in between each other in a consistency check. By harnessing additional correlation within the dataset the model's performance can be significantly increased compared to results from only structure from stereo or motion separately.
In future work the accuracy could benefit from developing a new sampler and modifying the explainability mask to include thin structures.

\ifCLASSOPTIONcaptionsoff
  \newpage
\fi



%
\bibliographystyle{IEEEtran}
\bibliography{egbib}

%








\end{document}